\documentclass[10pt,twocolumn,letterpaper]{article}

\usepackage{cvpr}              % To produce the CAMERA-READY version
\usepackage{balance}
\usepackage{makecell}
\usepackage{booktabs}
\usepackage{multirow}
\usepackage{minibox}
\usepackage{tabularx}
\usepackage{algorithmic}
\usepackage{lipsum}
\usepackage{amssymb} % for \checkmark symbol
\usepackage{pifont} % for ✗ symbol
\newcommand{\comment}[1]{}
\usepackage[ruled,vlined]{algorithm2e}
\usepackage{nicefrac}
\usepackage{svg}
\usepackage{subcaption}
\usepackage{amssymb,amsfonts}
\usepackage{dsfont}
\usepackage{xcolor}

\usepackage[accsupp]{axessibility} % Improves PDF readability for those with disabilities.

\usepackage{xcolor}

\definecolor{cvprblue}{rgb}{0.21,0.49,0.74}
\usepackage[pagebackref,breaklinks,colorlinks,citecolor=cvprblue]{hyperref}

\def\paperID{11} % *** Enter the Paper ID here
\def\confName{CVMI}
\def\confYear{2025}

\begin{document}

\title{SAM4EM: Efficient memory-based two stage prompt-free segment anything model adapter for complex 3D neuroscience electron microscopy stacks}

\author{Uzair Shah\\
CSE, HBKU\\
Doha, Qatar 
\and
Marco Agus\\
CSE, HBKU\\
Doha, Qatar\\
{\tt\small magus@hbku.equ.qa}
\and
Daniya Boges \\
CEMSE, KAUST \\
Thuwal, Saudi Arabia
\and
Vanessa Chiappini \\
University of Turin \\
Turin, Italy
\and
Mahmood Alzubaidi \\
CSE, HBKU \\
Doha, Qatar
\and
Jens Schneider \\
CSE, HBKU \\
Doha, Qatar 
\and
Markus Hadwiger \\
CEMSE, KAUST \\
Thuwal, Saudi Arabia
\and
Pierre J. Magistretti \\ 
BESE, KAUST \\
Thuwal, Saudi Arabia
\and 
Mowafa Househ \\
CSE, HBKU \\
Doha, Qatar 
\and
Corrado Cal\'i  \\
University of Turin \\
Turin, Italy \\
{\tt\small corrado.cali@unito.it}
}

\maketitle

%%%%%%%%% ABSTRACT
\begin{abstract}
  We present SAM4EM, a novel approach for 3D segmentation of complex neural structures in electron microscopy (EM) data by leveraging the Segment Anything Model (SAM) alongside advanced fine-tuning strategies. Our contributions include the development of a prompt-free adapter for SAM using two stage mask decoding to automatically generate prompt embeddings, a dual-stage fine-tuning method based on Low-Rank Adaptation (LoRA) for enhancing segmentation with limited annotated data, and a 3D memory attention mechanism to ensure segmentation consistency across 3D stacks.  We further release a unique benchmark dataset for the segmentation of astrocytic processes and synapses. We evaluated our method on challenging neuroscience segmentation benchmarks, specifically targeting mitochondria, glia, and synapses, with significant accuracy improvements over state-of-the-art (SOTA) methods, including recent SAM-based adapters developed for the medical domain and other vision transformer-based approaches. Experimental results indicate that our approach outperforms existing solutions in the segmentation of complex processes like glia and post-synaptic densities. Our code and models are available at \url{https://github.com/Uzshah/SAM4EM}.
\end{abstract}

%%%%%%%%% BODY TEXT
\section{Introduction}
\label{sec:intro}

\begin{figure}
    \centering
    \includegraphics[width=0.32\columnwidth]{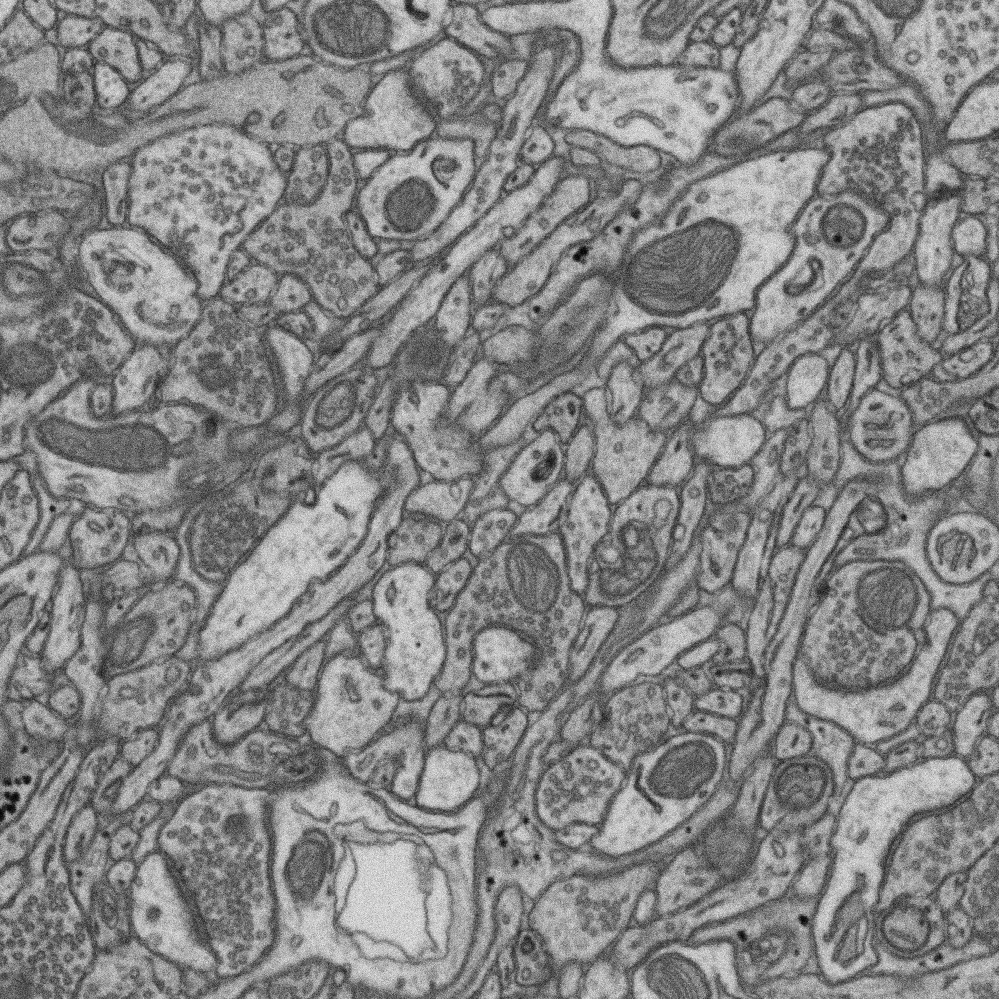}
    \includegraphics[width=0.32\columnwidth]{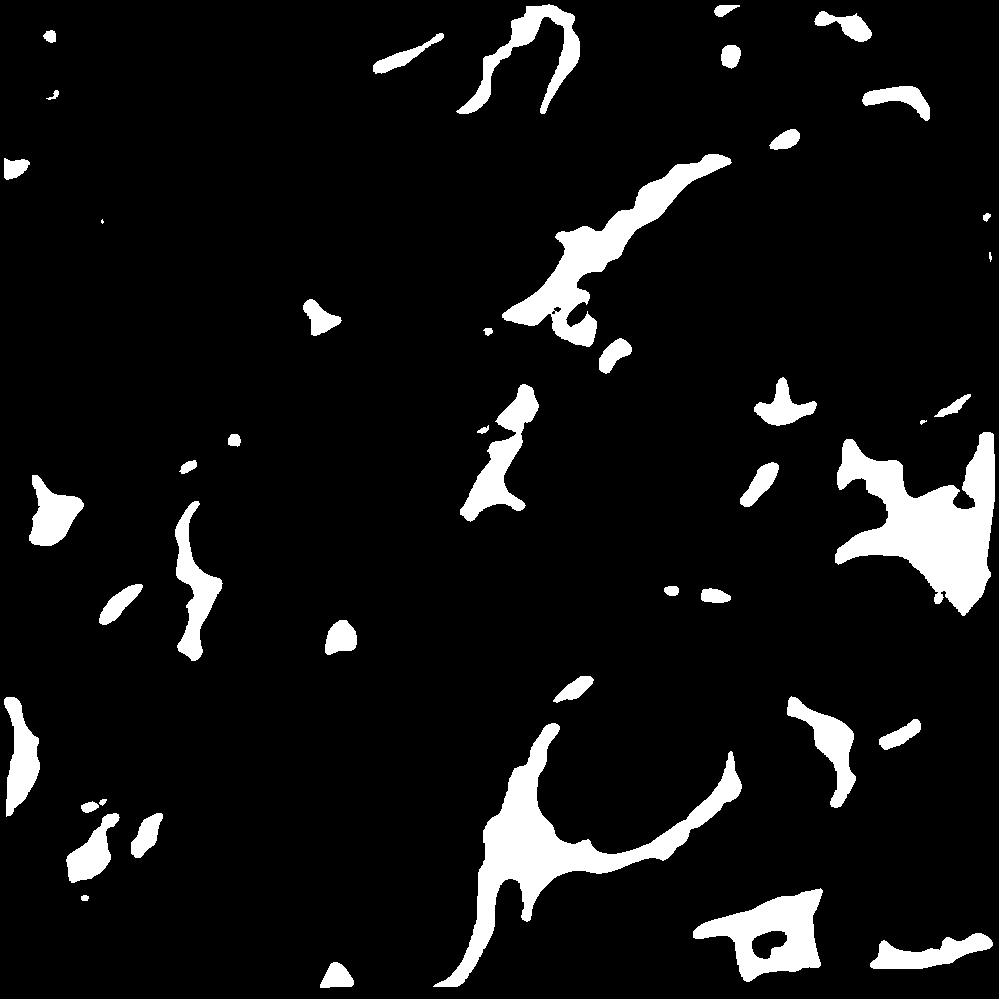}
    \includegraphics[width=0.32\columnwidth]{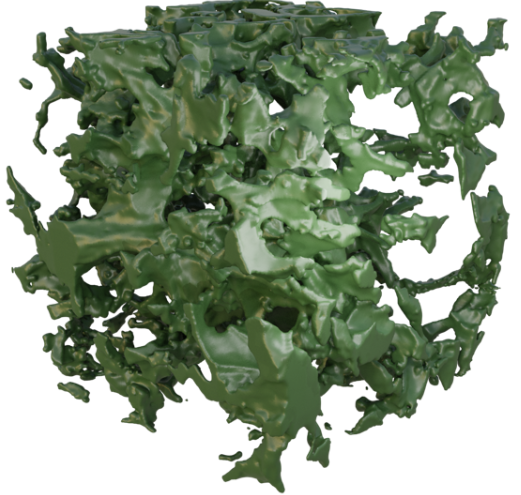}
    \caption{Current SOTA segmentation methods for 3D EM are effective for well-defined cellular structures like mitochondria~\cite{Luo:2024:frag}, while they struggle when segmenting complex processes with fractal structures like glia cells. We propose an efficient adapter for the Segment Anything Model which overcomes this limitation. Left: EM image. Center: glia annotation. Right: full glia 3D reconstruction.}
    \label{fig:teaser}
\end{figure}

Since the recent advent of electron microscopy imaging technologies, an increasing number of neuroscience institutes and practitioners began to include outcomes resulting from ultrastructural analysis of brain samples in their investigation pipeline~\cite{peddie:2022:volume}. The success of electron microscopy scanning technologies is due to the fact that they provide ways to clearly distinguish and recognize cellular and molecular structures at nanoscale resolutions, they enable accurate 3D reconstruction of neurons, and their organelles and surrounding cells. This all supports quantitative analysis and interactive visualization. 
In order to support fully ultrastructural investigations, accurate segmentation of the cells and organelles of interest is needed. This task traditionally involves time-consuming efforts from neuroscience experts and their collaborators~\cite{heinrich:2021:whole}. 
With the explosion of deep learning technologies, over the last five years, the computer vision community began to provide specific solutions for automatic segmentation of electron microscopy images. Common approaches are based on 
decoder-encoder convolutional neural networks~\cite{yuan:2021:hive-net}, and, more recently, on transformer architectures~\cite{Pan:2023:ATFormer,Luo:2024:frag}.
The proposed solutions provide promising results and started to be included inside neuroscience investigation pipelines~\cite{shapson-coe:2024:human,svara:2022:automated}. However, they are particularly data hungry in the worst sense. That is, they not only need large amounts of data, but specifically high-quality, annotated data. This is not easily available or feasible to generate in most labs.
Additionally, the most recently proposed techniques for automatic segmentation of 3D neural structures provide accurate results for 
cellular structures with well defined borders or with simple morphological shape, like mitochondria or nuclei. At the same time, they suffer in tracing cells exhibiting complex morphology processes, like glia or astrocytes~(see Fig.~\ref{fig:teaser}), or small structures with blurred borders,
like post-synaptic densities.

The recent release of the foundation Segment Anything Model~(SAM)~\cite{Kirillov:2023:SAM} is a potential game changer for the community. It has been trained on billions of masks and is potentially capable of targeting any kind of segmentation task, including the medical field~\cite{Huang:2024:SamMed}. 
The last version of SAM is also able to trace objects within videos. It has already been tested for automatic segmentation of organs in CT and MRI modalities~\cite{Huang:2024:SamMed}.
Following this trend, we examine the potential of SAM for segmentation of neuroscience electron microscopy stacks. We propose a novel effective end-to-end solution for fine-tuning with limited amount of annotated data for specific tasks related to neural structures with varying complexity, ranging from mitochondria, up to astrocytic processes, and post-synaptic densities.
In summary, we provide the following contributions:
\begin{itemize}
\item a prompt-free, iterative adapter for SAM that exploits hierarchical image embeddings and dual stage mask decoding to automatically generate prompt embeddings to fully automatize the segmentation process;
\item a dual stage fine-tuning strategy based on low rank adaptation~\cite{hu:2022:LoRA} which is able to refine the SAM decoder with limited amount of annotated data;  
\item a 3D memory attention mechanism on image and masks embedding, able to trace objects in 3D stacks, and enriching SAM with additional hints for maintaining consistency along the segmentation process;  
\item a novel benchmark dataset for Neuroscience EM containing 3D stacks and annotations regarding complex glia processes and sparse post-synaptic densities.
\end{itemize}

We evaluate the performance of our solution on challenging neuroscience segmentation tasks, specifically the labeling of 3D mitochondria organelles~(Lucchi~\cite{lucchi:2012:supervoxels} and Mito-EM~\cite{franco-barranco:2023:progress}), as well as
the reconstruction of complex astrocyte cells and post-synaptic densities on a novel curated dataset that we publicly release to the computer vision community. Up to our knowledge, this is the first publicly available annotated dataset for glia and synapse segmentation. Regarding the performances of the proposed SAM4EM method, our experiments show that our method is on par with the most recent, state-of-the-art solutions based on vision transformers~\cite{Pan:2023:ATFormer,Luo:2024:frag}, and on SAM adapters for medical images~\cite{Cheng:2024:H-SAM,Chen:2023:SAM-Adapter} for segmentation of mitochondria structures. At the same time, it is significantly outperforming current state-of-the-art solutions on more complex segmentation tasks, like astrocytic processes and synaptic densities. We also showcase the effects of our dual stage strategy, the low rank adaptation~(LoRA), and the 3D memory attention mechanism in dedicated ablation experiments. While LoRA has been proven successful for fine-tuning large language models~\cite{hu:2022:LoRA} and general vision transformers~\cite{yang:2023:LoRAvit,he:2023:pemavit}, we apply this approach in our SAM adaptation framework. Finally, to the best of our knowledge, this is the first solution adapting segment anything model to electron microscopy segmentation tasks.

%------------------------------------------------------------------------
\section{Related Work}
\label{sec:related}

Our work deals with automatic segmentation of electron microscopy images for neuroscience investigations and with parameter efficient fine-tuning of segmentation foundation models.
For the sake of brevity, we cannot extensively cover the literature related to these topics. In the following, we discuss the most recent techniques closely related to our method, while we refer interested readers to the recent surveys covering segmentation in large-scale cellular electron microscopy with deep learning~\cite{aswath:2023:survey}, transformer-based visual segmentation~\cite{li:2024:survey}, and application of segmentation foundation models to the biomedical field~\cite{Huang:2024:SamMed,Zhang:2024:SamMed}.

\paragraph*{Segmentation for electron microscopy}

Due to the laborious and time-consuming nature of manual segmentation of large-scale EM datasets, the computer vision community dedicates a lot of effort to the development of automated segmentation approaches. Recently, deep learning (DL) has emerged as the dominant approach in this field, and various solutions have been proposed for 2D and 3D semantic and instance segmentation. This was boosted by the public availability of high-quality annotated data as well as segmentation challenges~\cite{lucchi:2012:supervoxels,wei:2020:mitoem}. The first effective solutions followed the trends of other segmentation solutions, by adapting popular Full Convolutional Network architectures~(FCN)~\cite{fakhry:2016:res,mekuvc2022automatic}, like densely connected U-Net~\cite{cao:2020:densunet,quan:2021:fusionnet}, or recurrent networks with attention mechanisms
like the popular Flood Filling Network~\cite{januszewski2018high}. Very recently, various groups successfully applied vision transformers to the segmentation of neural organelles from electron microscopy stacks: ATFormer~\cite{Pan:2023:ATFormer} is a specific solution for mitochondria segmentation  based on appearance-adaptive templates for background, foreground,
and contour to sense the characteristics of different shapes of mitochondria, integrated in a hierarchical attention learning mechanism to absorb multi-level information. DualRel~\cite{Mai:2023:DualRel} consists of an end-to-end dual-reliable network. It is composed of a reliable pixel aggregation module and a reliable prototype selection module for  effective semi-supervised mitochondria segmentation. 
Liu et al.~\cite{Liu:2023:affcon} propose a sparsely supervised
instance segmentation framework. They consider  cross representation affinity consistency regularization and incorporate a pseudo-label noise filtering scheme as well as two entropy-based decision strategies. 
FragViT~\cite{Luo:2024:frag} is a coherent 3D vision transformer for mitochondria segmentation that interprets the 3D EM image stacks as a set of interrelated 3D fragments. It is based on a fragment encoder combined with affinity learning to manipulate features on 3D fragments and a hierarchical fragment aggregation module for progressively  aggregating fragment-wise predictions back to the final voxel-wise prediction. 
Finally, recent segmentation solutions have been tested on mitochondria segmentation benchmarks for generality: 
Distribution-Aware Weighting (DAW)~\cite{Sun:2023:DAW} consists of a semi-supervised semantic segmentation strategy  exploiting a large volume of unlabeled data to improve the model’s generalization
performance for robust segmentation. This is achieved by considering a weighting scheme to select pixel-level pseudo labels according to the confidence distribution of labeled and unlabeled data. RankMatch~\cite{Mai:2024:RankMatch} is a semi-supervised method based on the construction
of representative agents to model inter-pixel correlation beyond regular individual pixel-wise consistency, and on modeling inter-agent relationships in a way to achieve rank-aware correlation consistency. 

In contrast to previous methods, our solution consists of adapting a foundation model for segmentation through low-rank approximation to achieve prompt-free segmentation. We benchmark our method against state-of-the-art solutions proposed for mitochondria segmentation~\cite{Pan:2023:ATFormer,Mai:2023:DualRel,Luo:2024:frag}.

\paragraph*{Fine-tuning segmentation foundation models}
Segment Anything Model~\cite{Kirillov:2023:SAM} (SAM) is a foundation model trained on a huge segmentation dataset, comprising over 1 billion masks on 11 million images. SAM has been designed and trained to be promptable, in a way that it can transfer zero-shot to new image distributions and tasks. Immediately after the release of this foundation model, the community reacted with an impressive number of customizations, modifications. Since then, the model has been applied in many domains, including medicine and biology~\cite{Huang:2024:SamMed,Zhang:2024:SamMed}. However, depending on the task, zero-shot generalization does not provide the required segmentation accuracy. Therefore, various groups have recently developed strategies for efficient fine-tuning or adapting SAM for specific challenging tasks.
For example, SAM-Adapter~\cite{Chen:2023:SAM-Adapter} incorporates domain-specific information or visual prompts into the segmentation network by using simple yet effective adapters. EfficientSAM~\cite{Xiong:2024:EfficientSAM} leverages
masked image pre-training for constructing lightweight SAM models exhibiting decent performance with
largely reduced complexity. ASAM~\cite{Li:2024:ASAM} exploits the potential of natural adversarial examples by considering a stable diffusion model to  augment the SA-1B dataset with  adversarial examples that are more representative of natural variations. Peng et al.~\cite{Peng:2024:PeftSAM} propose a Parameter Efficient Fine Tuning~(PEFT) Scheme for SAM: they introduce an inter-block communication module, integrating a learnable relation matrix to facilitate communication among different coefficient sets of each PEFT block’s parameter space. HQ-SAM~\cite{Ke:2023:HQSAM} customizes SAM by integrating a learnable high-quality output token, which is injected into SAM’s mask decoder and fused to the mask-decoder's early and final ViT features for improved mask details.
Finally, H-SAM~\cite{Cheng:2024:H-SAM} is a prompt-free
adaptation of SAM tailored for efficient fine-tuning for medical images via a two-stage hierarchical decoding procedure.  SAM’s original decoder is used to generate an a-priori probabilistic mask, that guides a more intricate decoding process based on a class-balanced, mask-guided self-attention mechanism and a learnable mask cross-attention mechanism. 

Similarly to the latter, our solution consists of fine-tuning SAM by incorporating a dual stage low-rank adapter~\cite{hu:2022:LoRA} to obtain a prompt-free customized model. Moreover, we incorporate a memory attention mechanism similar to SAM2~\cite{ravi2024sam2} to enforce 3D consistency while tracing 3D objects.
To the best of our knowledge, this is the first attempt of customizing SAM for electron microscopy segmentation tasks.

\section{Background}

The Segment Anything Model (SAM) is a foundation model for image segmentation. SAM is trained on a large dataset of natural images. SAM's architecture consists of three primary components: an image encoder, a prompt encoder, and a lightweight mask decoder. The image encoder---based on the Vision Transformer (ViT) architecture---processes the input image to generate vision embeddings. The prompt encoder, utilizing positional encoding summed with learned embeddings for each prompt type, accepts points, boxes, and masks. The latter are embedded using convolutions and summed with the vision embedding. The mask decoder, which employs a bidirectional transformer decoder and a pixel decoder, processes image embeddings from the image encoder along with prompt embeddings. The transformer decoder uses self-attention to assess the significance of different image regions and cross-attention to focus on relevant areas for segmentation. The pixel decoder further refines these outputs. In summary, SAM takes an input image and a prompt to generate a segmentation mask.

Very recently, Meta released SAM2~\cite{ravi2024sam2}, an advanced version of SAM trained on both image and video data, offering more power and efficiency. SAM2 maintains the same structure as SAM, with an image encoder that generates embeddings from input images and a lightweight mask decoder that produces the segmentation output based on various prompts (points, boxes, or masks). In addition, SAM2 also introduces a memory encoder, memory attention, and a memory bank for video processing. Unlike SAM, SAM2 conditions the frame embeddings not directly from the image encoder but on memories of past predictions and prompted frames, including those from the future relative to the current frame. The memory encoder creates frame memories based on the current prediction, storing them in a memory bank for future use. The memory attention mechanism then conditions the per-frame embedding from the image encoder on this memory bank, which is subsequently passed to the mask decoder.

\section{SAM4EM Architecture}
\subsection{Overview}
\begin{figure*}
\centering
\includegraphics[width=1\linewidth]{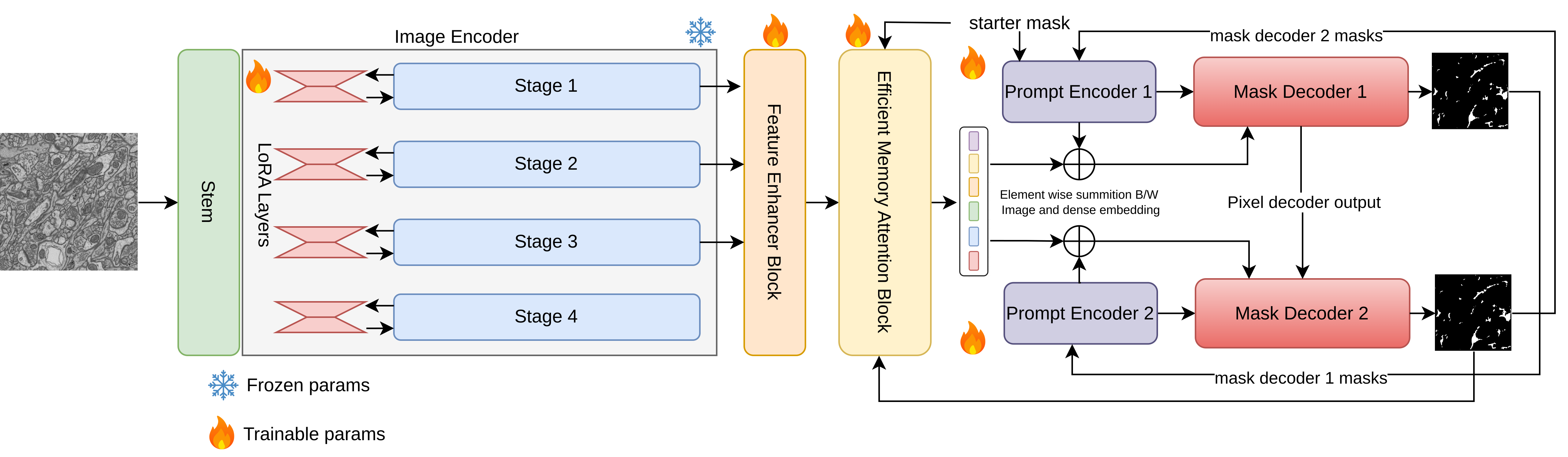}
\caption{Overview of the proposed SAM4EM architecture. Our model extends SAM with three key components: (a) feature enhancer block (Sec.~\ref{sec:enhancer}), (b) memory-based encoder (Sec.~\ref{sec:memory}), and (c) bi-directional prompt mechanism (Sec.~\ref{sec:prompt}), integrated with residual connections for accurate medical image segmentation.}
\label{fig:arch}
\end{figure*}

Our SAM4EM architecture, built upon SAM2, comprises five key components: a LoRA-adapter image encoder (Sec. \ref{sec:LoRA}, a feature enhancer block (Sec. \ref{sec:enhancer}), an efficient memory encoder block (Sec. \ref{sec:memory}), a bi-directional self-prompting module (Sec. \ref{sec:prompt}). Figure~\ref{fig:arch} provides an illustration of our architecture. Our SAM4EM system processes multi-resolution features through a feature enhancer block with fusion layers, while incorporating previous slice mask outputs to condition the current slice via the memory encoder block. The architecture employs a dual-stage decoding process: the first prompt encoder generates dense and sparse embeddings from previous slice masks, which combine with memory-conditioned image features to produce initial mask hints and attention maps. These outputs are subsequently refined through a second prompt encoder and decoder stage to generate the final masks. This self-prompting mechanism, coupled with memory-based feature propagation along the z-axis of EM stacks, enables efficient slice-by-slice segmentation processing.
% Given an image \( I \) of size \( W \times H \), the goal is to predict its segmentation map, where each pixel is assigned to a predefined category, closely matching the ground truth. Our framework, SAM4EM, is closely aligned with H-SAM \cite{Cheng:2024:H-SAM} and builds upon SAM2, featuring a LoRA-adapted image encoder and a 2-stage hierarchical mask decoder. In the first stage, the decoder generates a segmentation map using image embeddings and default prompts. This map is then used as a prompt for the second decoder. We calculate spatial attention between the image embedding and the probability maps from the first decoder, yielding a refined image embedding. This refined embedding is combined with the prompt and passed to the second decoder, resulting in a final segmentation map.

\subsection{LoRA-adapted Image Encoder}
\label{sec:LoRA}
To maintain the pre-trained knowledge whilst enabling task-specific adaptation, SAM4EM implements a Low-Rank Adaptation (LoRA) strategy in its image encoder. The original SAM encoder layers remain frozen, while parallel trainable bypass paths are introduced. These bypass paths consist of paired low-rank matrices that first project the transformer features into a compressed representation before expanding them to match the original feature dimensionality. This approach minimizes the number of trainable parameters while allowing effective model fine-tuning through the optimization of only the bypass matrices, achieving a balance between preserving learned representations and adapting to specialized tasks.
\subsection{Feature Enhancer Block}
\label{sec:enhancer}
\begin{figure}
\centering
\includegraphics[width=1\linewidth]{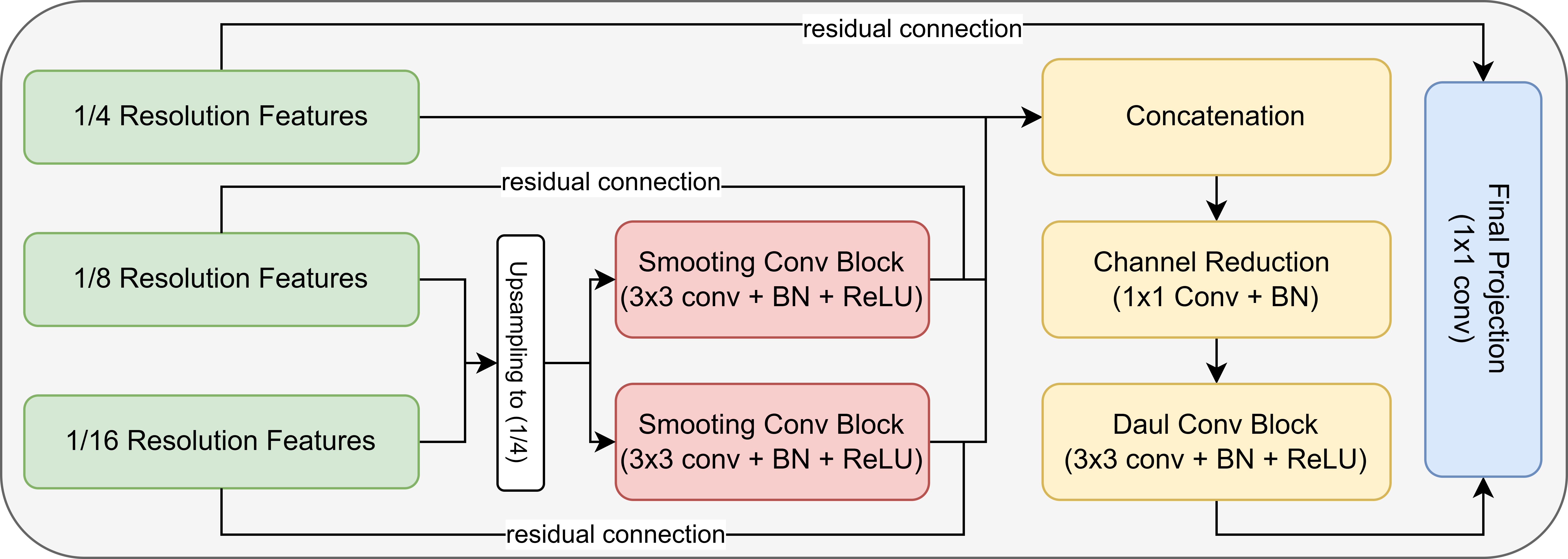}
\caption{Architecture of the proposed feature enhancer block. Multi-resolution features ($\nicefrac{1}{4}$, $\nicefrac{1}{8}$, $\nicefrac{1}{16}$) are processed through parallel pathways with up-sampling, smoothing convolutions, and residual connections. Features are fused via concatenation and refined through channel reduction and dual convolution blocks for precise boundary delineation.}
\label{fig:feat_blk}
\end{figure}

SAM2's default approach of generating quarter-resolution masks with bilinear interpolation proves inadequate for medical image segmentation, where irregular anatomical shapes and high-frequency details demand precise boundary delineation. To address this limitation, SAM4EM introduces a novel feature enhancer block that leverages multi-scale processing ($\nicefrac{1}{4}$, $\nicefrac{1}{8}$, and $\nicefrac{1}{16}$ resolutions), enabling simultaneous capture of fine-grained details and broader anatomical context as shown in Figure~\ref{fig:feat_blk}.

The block implements a hierarchical enhancement strategy where $\nicefrac{1}{8}$ and $\nicefrac{1}{16}$ resolution features are upsampled through a smoothing convolution block ($3\times3$ convolution, batch normalization, ReLU activation) with residual connections. These residual pathways maintain direct gradient flow and preserve essential low-level anatomical details. The enhanced features undergo fusion through a sophisticated pipeline: concatenation, followed by a $1\times1$ convolution for channel reduction, batch normalization, and dual $3\times3$ convolution blocks. A final residual connection to the initial $\nicefrac{1}{4}$ resolution feature, coupled with a $1\times1$ projection layer, ensures preservation of fine anatomical details while integrating contextual information.
This comprehensive architecture eliminates the need for interpolation as a post-processing step and directly generates high-fidelity segmentation masks that accurately capture intricate medical structures. The multi-scale approach particularly excels in delineating complex anatomical boundaries and accommodating varying structural scales, significantly improving segmentation accuracy compared to traditional single-scale approaches.

\subsection{Efficient Memory Encoder}
\label{sec:memory}
While SAM2's memory encoder effectively handles temporal relationships, it faces significant limitations in medical image analysis: computational overhead for long image sequences, difficulty adapting to morphological variations across consecutive slices, and incompatibility with high-resolution features. Additionally, it often requires manual intervention when new anatomical structures appear in later stages. To address these limitations, we propose an efficient alternative to SAM2's memory encoder and attention mechanism that maintains performance while reducing computational complexity.

Our proposed efficient memory encoder implements a lightweight attention mechanism that effectively balances temporal information integration with computational efficiency. The system processes feature maps of size $128\times128$ with 256 channels through dimension reduction, projecting them into a more compact 128-channel memory space to minimize computational overhead. A key innovation is the dual-stream processing approach that combines current slice features with previous mask information. This integration is achieved through a lightweight convolutional encoder, consisting of two strided convolutions with ReLU activations, which processes the previous mask and aligns it spatially with current features through interpolation. The additive combination of encoded mask features and projected current slice features creates a comprehensive representation that captures both spatial and temporal contexts efficiently.

The attention mechanism employs an optimized dot-product formulation. For a query $\mathbf{Q}$ derived from the combined feature-mask representation and key-value pairs $\left(\mathbf{K}, \mathbf{V}\right)$ from the memory slots $(M=8)$, the attention is computed as:
\begin{align}
\mathbf{A} &= \text{softmax}\left(\frac{\mathbf{Q}\mathbf{K}^T}{\sqrt{d}}\right)\mathbf{V}
\end{align}
where $d$ represents the memory dimension (128), and the scaling factor $\sqrt{d}$ prevents gradient vanishing. This formulation preserves spatial dimensions throughout the computation, maintaining precise location-specific feature correspondences across slices.

The memory update mechanism utilizes an exponential moving average with a momentum parameter $\alpha$:
\begin{align}
\mathbf{M}_t &= (1-\alpha)\mathbf{M}_{t-1} + \alpha\mathbf{F}_t
\end{align}
where $\mathbf{M}_t$ denotes the current memory state, $\mathbf{M}_{t-1}$ the previous state, and $\mathbf{F}_t$ the combined feature-mask representation. This approach ensures smooth temporal feature propagation while adaptively incorporating new slice characteristics, effectively preventing segmentation inconsistencies. The conservative momentum value emphasizes historical information, maintaining robust temporal consistency while allowing gradual adaptation to anatomical variations.

\subsection{Bi-directional Self-Prompting}
\label{sec:prompt}
\noindent\textbf{Prompt Encoder} Our implementation builds upon the original SAM prompt encoder, with modifications to accommodate high-resolution image embeddings through updated positional embeddings. Through extensive experimentation, we discovered that mask-based prompts, despite inherent noise, consistently outperform traditional point or box prompts in segmentation accuracy. This insight led to the development of our novel self-prompting loop mechanism, which implements sequential mask refinement through a dual-stage encoding process.

The architecture features two synchronized prompt encoders working in complementary roles. The first encoder processes the previous slice's segmentation output as an initial noisy prompt, serving dual purposes: maintaining temporal consistency for existing structures while adapting to anatomical variations. This initial encoding generates preliminary mask hints through the first mask decoder, which are subsequently refined by the second prompt encoder. This hierarchical structure proves particularly effective for medical image segmentation, where anatomical structures frequently evolve between consecutive slices.

The dual-encoder approach effectively addresses key challenges in volumetric segmentation. The initial encoder-decoder pair functions acts as a coarse segmentation stage, processing both persistent and transient structures while maintaining robustness to noise from previous slice outputs. The second encoder then utilizes this initial mask to generate refined prompt embeddings, effectively reducing noise and enhancing segmentation precision. This creates an iterative refinement loop that benefits from both temporal context and local feature enhancement.

Our self-prompting mechanism establishes bi-directional information flow: the first prompt encoder utilizes the final output from the second mask decoder, while the second prompt encoder refines the initial output from the first mask decoder. This recursive architecture enables robust temporal consistency while adapting to slice-specific variations.\\[-0.5em]

\noindent\textbf{Mask Decoder} SAM4EM implements a sophisticated two-stage hierarchical decoding procedure. The first stage utilizes a modified version of SAM's original decoder to generate a prior mask, which guides subsequent refined decoding. While both decoders maintain similarity with the original SAM architecture, the second decoder incorporates specific enhancements for improved performance. Drawing inspiration from H-SAM, we implement a hierarchical pixel decoder in the second stage that integrates features from the first-stage decoder through skip connections, enabling high-resolution prediction generation with enhanced detail preservation.\\[-0.5em]

This comprehensive prompt encoding and mask decoding framework ensures robust volumetric segmentation while maintaining fine anatomical details across consecutive slices.

\begin{table*}
\centering
\caption{Quantitative comparison across datasets using Dice coefficient (Dice$\uparrow$) and mean Intersection over Union (mIoU$\uparrow$) metrics. Higher values indicate better performance. Our method consistently outperforms baseline approaches across all datasets.}
\begin{tabular}{l|c|c|c|c|c|c|c|c}
\hline
\multirow{2}{*}{\textbf{Model}} & \multicolumn{2}{c|}{\textbf{Mice-Glia}} & \multicolumn{2}{c|}{\textbf{Mice-Mito}} & \multicolumn{2}{c|}{\textbf{Mice-Synapses}} & 
 \multicolumn{2}{c}{\textbf{Lucchi}}  \\
\cline{2-9}
& Dice$\uparrow$ & mIoU$\uparrow$ & Dice$\uparrow$ & mIoU$\uparrow$ & Dice$\uparrow$ & mIoU$\uparrow$ & Dice$\uparrow$ & mIoU$\uparrow$ \\
\hline
H-SAM \cite{Cheng:2024:H-SAM} & 68.7 & 53.6 & 74.3 & 60.2 
& 42.3 & 27.6 & 90.5 & 83.2 \\
SAMed \cite{zhang2023customized} & 66.9 & 51.7 & 71.5 & 56.9 & 38.5 & 25.1 & 88.4 & 79.8  \\
UN-SAM \cite{chen2024sam} & 54.3 & 37.6 &  60.8  &  44.4& 40.2 & 26.3 & 87.2 & 77.5 \\
SAM4EM & \textbf{70.5} & \textbf{54.9} & \textbf{80.7} & \textbf{69.1} & \textbf{53.8} & \textbf{37.8} & \textbf{92.4} & \textbf{86.1} \\
\hline
\end{tabular}
\label{tab:model-comparison}
\end{table*}

\begin{figure}[t]
\centering
\includegraphics[width=\linewidth]{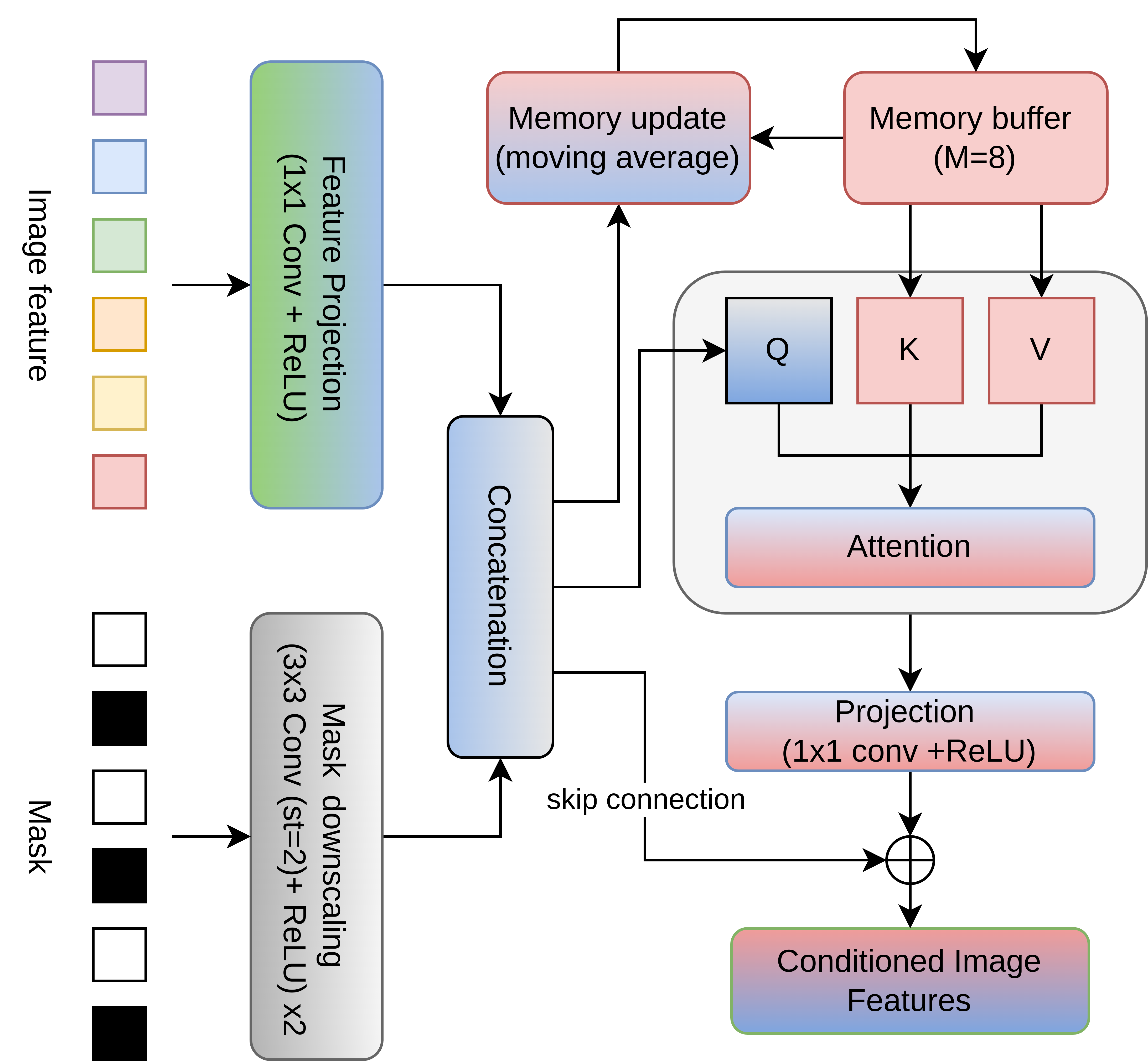}
\caption{Architecture of the efficient memory encoder. Image and mask features are processed through parallel streams with feature projection and mask downsampling, then integrated with a memory buffer (M=8) via attention mechanism. Moving average updates and skip connections maintain temporal consistency for feature conditioning.}
\label{fig:memory_encoder}
\end{figure}

\section{Experimental Setup}
\paragraph{Implementation details}
We implemented SAM4EM using the PyTorch framework. We conducted training on an NVIDIA RTX2080 GPU. We initialized the model with SAM2-base weights and compared against state-of-the-art medical image segmentation models, specifically H-SAM and SAMed ViT-B. Training was optimized using AdamW with an initial learning rate of $5\times 10^{-4}$ and ran for 200 epochs to ensure convergence.

All images were pre-sampled to a standardized resolution of $512\times512$ pixels. To enhance model robustness and prevent overfitting, we employed comprehensive data augmentation techniques, including random rotation, scaling, elastic deformation, flipping, and intensity enhancement. The rationale for these particular augmentation operations is that these parameters are hard to control in an EM setting: tissue slices vary in thickness, resulting in intensity variations; they are hard to place and orient under the EM; and they are soft, resulting in elastic deformations and stretching.

Our implementation uses LoRA adaptation to reduce trainable parameters by ~85\%, requiring only 4GB of GPU memory during training. The model processes large datasets efficiently via sliding window inference, enabling application to volumes significantly larger than training data.

The model's performance was optimized through random grid search over hyperparameters, resulting in an optimal memory momentum parameter $\alpha$ of 0.3 and 8 memory slots for the efficient memory encoder.

The loss function was formulated as a weighted combination of Dice loss and binary cross-entropy, with equal weights (0.5) assigned to each component:

\begin{align}
\mathcal{L}_{\text{total}} &= \frac{1}{2}\mathcal{L}_{\text{Dice}} + \frac{1}{2}\mathcal{L}_{\text{bce}}
\end{align}

\paragraph*{Datasets}
\begin{itemize}
\item \textbf{Lucchi dataset}~\cite{lucchi:2012:supervoxels}: An isotropic focussed Ion beam scanning electron microscope (FIB-SEM) volume imaged from the hippocampus of a mouse brain. It has the same spatial resolution along all three axes. This dataset has now become the de-facto standard for evaluating mitochondria segmentation performance. 
% An enhanced version of this benchmark dataset, the Lucchi++ dataset, was presented by 
% Casser et al.~\cite{casser:2020:mito},  with re-annotations that ensured consistent mitochondria boundaries and corrections of misclassifications.
% \item \textbf{Mito-EM}~\cite{wei:2020:mitoem} it is the largest mammalian mitochondria dataset from humans (MitoEM-H) and adult rats (MitoEM-R). It is about 3600 times larger than the Lucchi dataset described below, which has become a standard dataset for mitochondria segmentation and contains mitochondria instances of at least 2000 voxels in size. Complex morphology such as mitochondria on a string (MOAS) connected by thin microtubules or instances entangled in 3D were captured using ssSEM. The MitoEM dataset was created to provide a comprehensive view of the ultrastructure of mitochondria and to facilitate a comparative study of mitochondrial morphology and function in rats and humans.
\item \textbf{Mices- dataset}: we curated a novel dataset from the EM stacks used for a study about the effects of aging in glycogen distribution~\cite{cali2018effects}. To this end we extracted the semantic signals related to glia, synapsis and mitochondria on 3D stacks representing the  somatosensory cortex of adult and aged mice. The stack has size of 5 $\mu m^3$, and it is composed by 448 slices at 1000 x 1000 resolution. Additional details about the novel released dataset can be found in the supplementary materials.
\end{itemize}

% \paragraph{Training strategy}

\section{Results}
\label{sec:results}
Our experimental evaluation reveals compelling evidence for the effectiveness of our proposed model across diverse biomedical image segmentation tasks. Through comprehensive comparisons with other prompt-free models, we demonstrate consistent performance improvements across multiple datasets.\\[-0.6em]

\noindent\textbf{Mice-Glia dataset.}
For glia cell segmentation, our model achieves remarkable results with a Dice score of 70.5\% and mean IoU of 54.9\%. As shown in Table~\ref{tab:model-comparison}, these metrics surpass both the H-SAM architecture~\cite{Cheng:2024:H-SAM} (68.7\% Dice, 53.6\% mIoU) and SAMed~\cite{zhang2023customized} (66.9\% Dice, 51.7\% mIoU). The improvement of 1.8 percentage points in Dice score over the state-of-the-art H-SAM model highlights our method's enhanced capability in precise glia cell boundary delineation.\\[-0.6em]

\noindent\textbf{Mice-Mito dataset.}
In the challenging domain of mitochondria segmentation, our approach demonstrates robust performance, achieving a Dice score of 80.7\% and mIoU of 69.1\%. Table~\ref{tab:model-comparison} shows that these results compare favorably against H-SAM~\cite{Cheng:2024:H-SAM} (74.3\% Dice, 60.2\% mIoU) and SAMed~\cite{zhang2023customized} (71.5\% Dice, 56.9\% mIoU). The enhanced performance can be attributed to our model's ability to capture the intricate morphological features characteristic of mitochondrial structures.\\[-0.6em]

\noindent\textbf{Mice-Synapses dataset.}
Perhaps the most striking improvements we observed are in synaptic structure segmentation, where our model achieved a Dice score of 53.8\% and mIoU of 37.8\%. These results represent substantial advances over both H-SAM~\cite{Cheng:2024:H-SAM} (42.3\% Dice, 27.6\% mIoU) and SAMed~\cite{zhang2023customized} (38.5\% Dice, 25.1\% mIoU). The remarkable gain of 11.5 percentage points in Dice score compared to H-SAM underscores our model's particular strength in handling the complexities of synaptic junction detection.\\[-0.6em]

\noindent\textbf{Lucchi dataset.}
To further validate the performance of our proposed method, we used the publicly available Lucchi dataset. Our model exhibited particularly impressive results, achieving a Dice score of 86.1\% and an exceptional mIoU of 92.4\%. These metrics represent a meaningful improvement over both H-SAM~\cite{Cheng:2024:H-SAM} (90.5\% Dice, 83.2\% mIoU) and SAMed~\cite{zhang2023customized} (88.4\% Dice, 79.8\% mIoU), particularly in terms of mIoU performance. The notably high mIoU score indicates our model's superior capability in precise cellular structure delineation.\\[-0.6em]

\noindent\textbf{Overall performance.}
The consistent pattern of improvement across these diverse datasets speaks to the robustness and versatility of our approach. While UN-SAM results were unavailable for comparison, our model's strong performance across various cellular and subcellular segmentation tasks demonstrates its potential as a reliable tool for biomedical image analysis. These results collectively suggest that our architectural innovations successfully address key challenges in medical image segmentation while maintaining broad applicability across different biological contexts.\\[-0.6em]

\begin{figure}[htb!]
    \centering
    \includegraphics[width=\columnwidth]{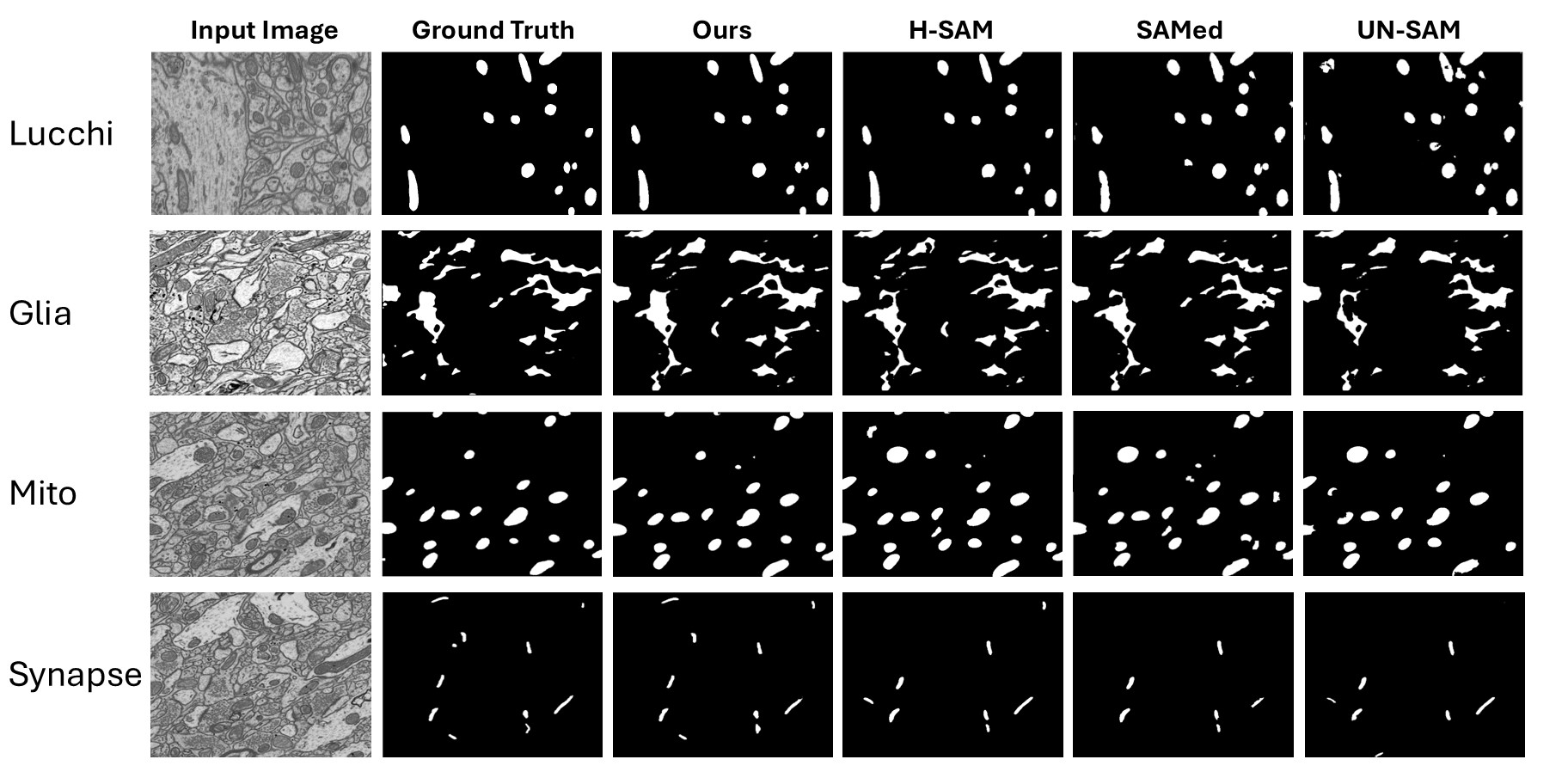}
    \caption{Qualitative comparison of segmentation results across different datasets and methods. From left to right: input electron microscopy images, ground truth masks, and segmentation results from our proposed method, H-SAM, SAMed, and UN-SAM. The comparison spans four different datasets: Lucchi (mitochondria), Glia (glial cells), Mito (mitochondria), and Synapse (synaptic junctions). Our method demonstrates superior boundary delineation and structural preservation across varied anatomical structures, particularly evident in the complex morphologies of glial cells and the fine details of synaptic junctions.}
    \label{fig:qualitative}
\end{figure}

\noindent\textbf{Qualitative results.}
Figure~\ref{fig:qualitative} presents a visual comparison of segmentation results across different methods on four challenging electron microscopy datasets. Our model demonstrates superior performance in preserving fine structural details and accurate boundary delineation. On the Lucchi dataset, our method accurately segments mitochondria with clear boundaries, while other methods show slight inconsistencies. In the Glia dataset, our approach better captures the complex, irregular morphology of glial cells, maintaining continuity in branched structures where competing methods show fragmentation. For the Mito dataset, our method achieves more precise mitochondrial boundary detection with fewer artifacts compared to baseline approaches. Most notably, in the challenging Synapse dataset, our model effectively identifies and segments fine synaptic junctions while maintaining structural integrity, whereas other methods struggle with these delicate features. These visual results complement our quantitative findings and highlight our model's capability to handle diverse anatomical structures across different imaging conditions.

\paragraph*{Ablation experiments.}
To better understand the contribution of each architectural component, we conducted a comprehensive ablation study on the Lucchi dataset, focusing on three key innovations: Feature enhancement, memory encoder, and bi-directional prompt mechanism. All experiments maintained default hyperparameters and ran for 20 epochs to ensure fair comparison. 
Table~\ref{tab:ablation} presents the results of our ablation study. Our baseline model achieved 75.3\% mIoU. Testing individual components revealed interesting patterns: The memory encoder alone improved performance to 78.5\%, highlighting its strength in temporal information processing. The bi-directional prompt mechanism reached 77.2\% mIoU, demonstrating effective boundary refinement. Feature enhancement proved especially powerful, achieving 78.3\% mIoU through better feature representation.
Component combinations yielded even more compelling results. Feature Enhancement paired with the memory encoder showed particularly strong synergy, reaching 81.3\% mIoU. Other combinations also performed well: pairing the memory encoder with bi-directional prompts achieved 80.1\%, while feature enhancement with bi-directional prompt reached 78.4\%.
The full model, incorporating all three components, achieved our best result of 82.3\% mIoU---a significant 7-point improvement over the baseline. This substantial gain demonstrates how each component addresses different aspects of the segmentation challenge while working together harmoniously. These results validate our architectural choices and show that even with a modest 20-epoch training regime, our approach effectively tackles the complexities of medical image segmentation.

\begin{table}[htb!]
\centering
\caption{Ablation study results showing mIoU performance with different component combinations. Checkmarks (\checkmark) indicate enabled components, crosses (\ding{55}) indicate disabled ones. Full model with all components achieves best mIoU of 82.3\%.}
\begin{tabular}{c|c|c|c}
\hline
Feat. Enh. & Mem. Encoder & Bi-dir. Prompt & mIoU$\uparrow$ \\
\hline
\ding{55} & \ding{55} & \ding{55} & 75.3 \\
\ding{55} & \checkmark & \ding{55} & 78.5 \\
\ding{55} & \ding{55} & \checkmark & 77.2 \\
\ding{55} & \checkmark & \checkmark & 80.1 \\
\checkmark & \ding{55} & \ding{55} & 78.3 \\
\checkmark & \checkmark & \ding{55} & 81.3 \\
\checkmark & \ding{55} & \checkmark & 78.4 \\
\checkmark & \checkmark & \checkmark & 82.3 \\
\hline
\end{tabular}
\label{tab:ablation}
\end{table}

\section{Conclusion}
\label{sec:conclusion}

In this work, we introduced SAM4EM, an efficient, two-stage prompt-free adapter for the Segment Anything Model (SAM) tailored to the unique requirements of 3D electron microscopy segmentation in neuroscience. By leveraging a low-rank adaptation (LoRA) image encoder, a novel feature enhancer block, and an efficient memory encoder, our approach  balances high segmentation accuracy with computational efficiency. Additionally, the bi-directional self-prompting mechanism enables robust segmentation across consecutive slices, ensuring temporal consistency and refined boundary delineation for complex cellular structures.
Our extensive evaluations across multiple challenging neuroscience segmentation datasets, including mitochondria, glia, and synapse segmentation, demonstrate that SAM4EM  outperforms existing state-of-the-art methods in both accuracy and consistency metrics. Ablation studies further validate the contributions of each architectural component, showcasing their collective impact on segmentation quality. As future work, we plane to extend the method to deal with dense segmentation tasks involving the recognition of cellular structures and organelles at various scales and with different taxonomies.

{\small
\textbf{Acknowledgments.}
This publication was funded by the PPM-7th Cycle grant (PPM 07-0409-240041, AMAL-For-Qatar) from the Qatar National Research Fund, a member of the Qatar Foundation. The findings herein reflect the
work and are solely the responsibility, of the authors.
}

%%%%%%%%% REFERENCES
\clearpage
{\small
\bibliographystyle{ieee_fullname}
\bibliography{sam-em}
}
 
\end{document}

\title{SAM4EM: efficient memory-based two stage prompt-free segment anything model adapter for complex 3D neuroscience electron microscopy stacks}

\author{Uzair Shah\\
CSE, HBKU\\
Doha, Qatar 
\and
Marco Agus\\
CSE, HBKU\\
Doha, Qatar\\
{\tt\small magus@hbku.equ.qa}
\and
Daniya Boges \\
CEMSE, KAUST \\
Thuwal, Saudi Arabia
\and
Vanessa Chappini \\
University of Turin \\
Turin, Italy
\and
Mahmood Alzubaidi \\
CSE, HBKU \\
Doha, Qatar
\and
Jens Schneider \\
CSE, HBKU \\
Doha, Qatar 
\and
Markus Hadwiger \\
CEMSE, KAUST \\
Thuwal, Saudi Arabia
\and
Pierre J. Magistretti \\ 
BESE, KAUST \\
Thuwal, Saudi Arabia
\and 
Mowafa Househ \\
CSE, HBKU \\
Doha, Qatar 
\and
Corrado Cal\'i  \\
University of Turin \\
Turin, Italy \\
{\tt\small corrado.cali@unito.it}
}
\maketitle

%%%%%%%%% ABSTRACT
\begin{abstract}
  We present SAM4EM, a novel approach for 3D segmentation of complex neural structures in electron microscopy (EM) data by leveraging the Segment Anything Model (SAM) alongside advanced fine-tuning strategies. Our contributions include the development of a prompt-free adapter for SAM using two stage mask decoding to automatically generate prompt embeddings, a dual-stage fine-tuning method based on Low-Rank Adaptation (LoRA) for enhancing segmentation with limited annotated data, and a 3D memory attention mechanism to ensure segmentation consistency across 3D stacks.  We further release a unique benchmark dataset for the segmentation of astrocytic processes and synapses. We evaluated our method on challenging neuroscience segmentation benchmarks, specifically targeting mitochondria, glia, and synapses, with significant accuracy improvements over state-of-the-art (SOTA) methods, including recent SAM-based adapters developed for the medical domain and other vision transformer-based approaches. Experimental results indicate that our approach outperforms existing solutions in the segmentation of complex processes like glia and post-synaptic densities. Our code and models are available at \url{https://github.com/Uzshah/SAM4EM}.
\end{abstract}

\section{Introduction}

In the following we provide further information about the process we carried out for annotating the data~(Sec.~\ref{sec:data}), and additional benchmarks against the most recent SOTA transformer-based methods~(Sec.~\ref{sec:vit}), including a qualitative analysis on noisy microscope data performed by domain scientists participating to the project~(Sec.~\ref{sec:qual}).

\section{Data Curation}
\label{sec:data}

\begin{figure*}
    \centering
    \includegraphics[width=0.32\textwidth]{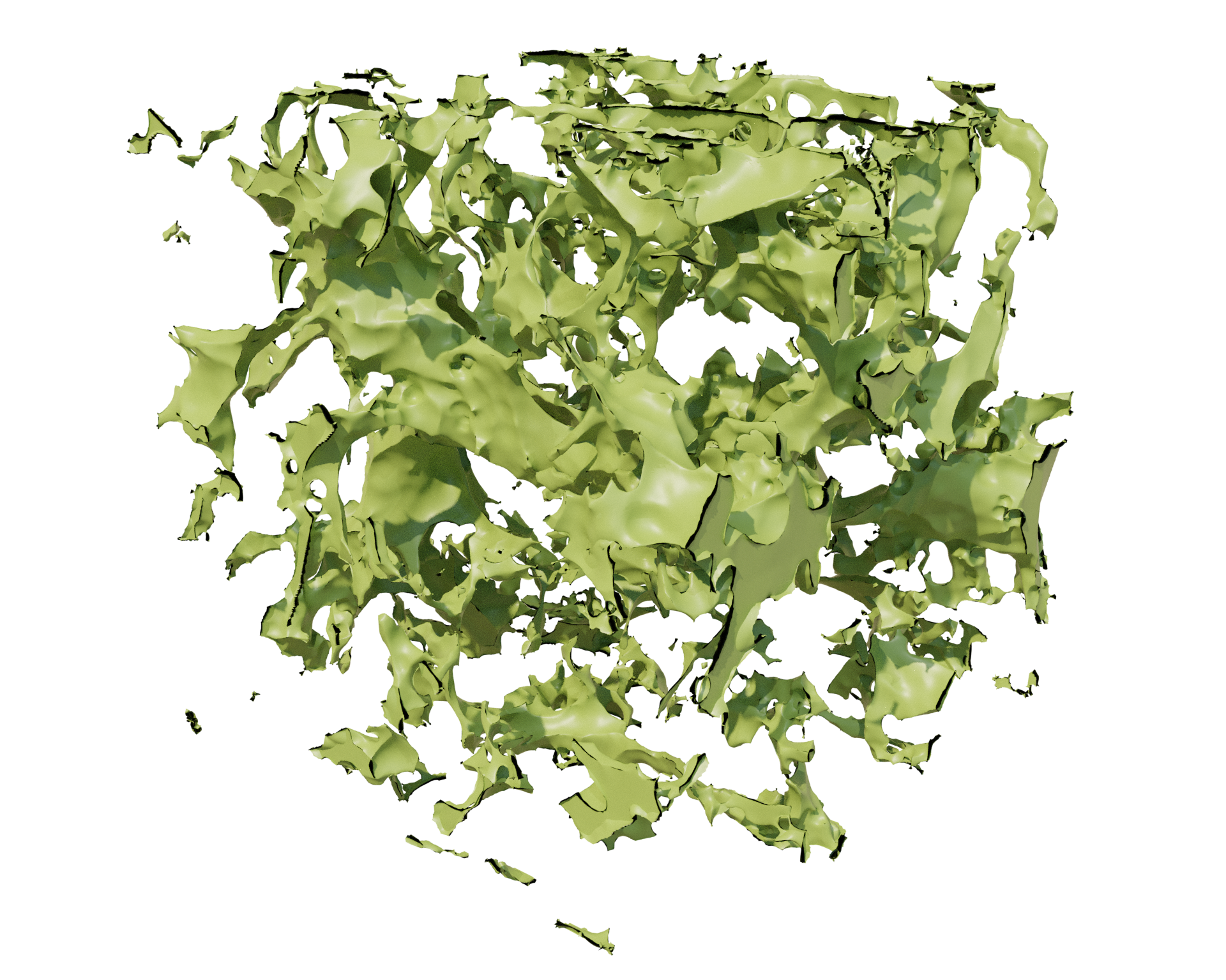}
    \includegraphics[width=0.32\textwidth]{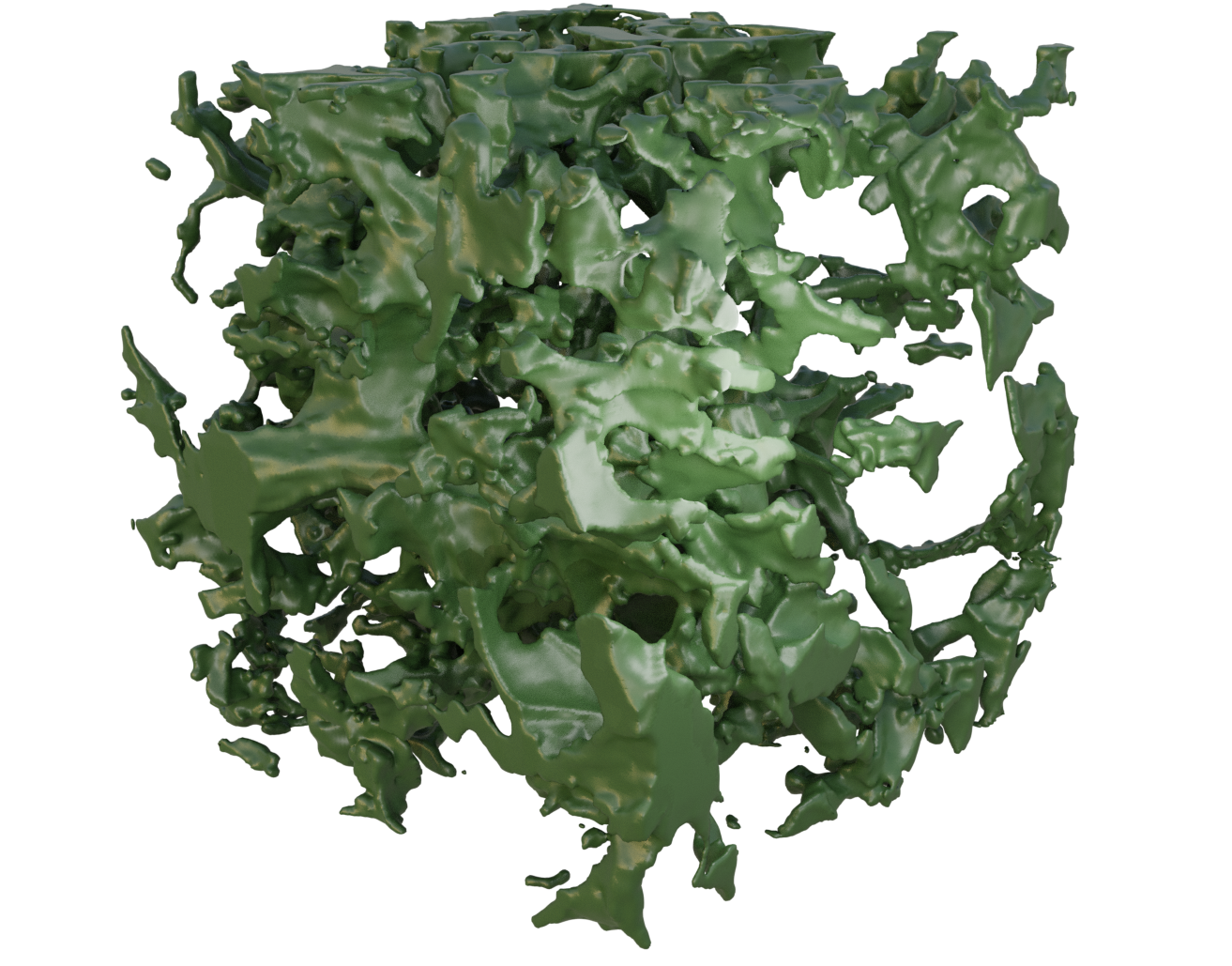}
    \includegraphics[width=0.32\textwidth]{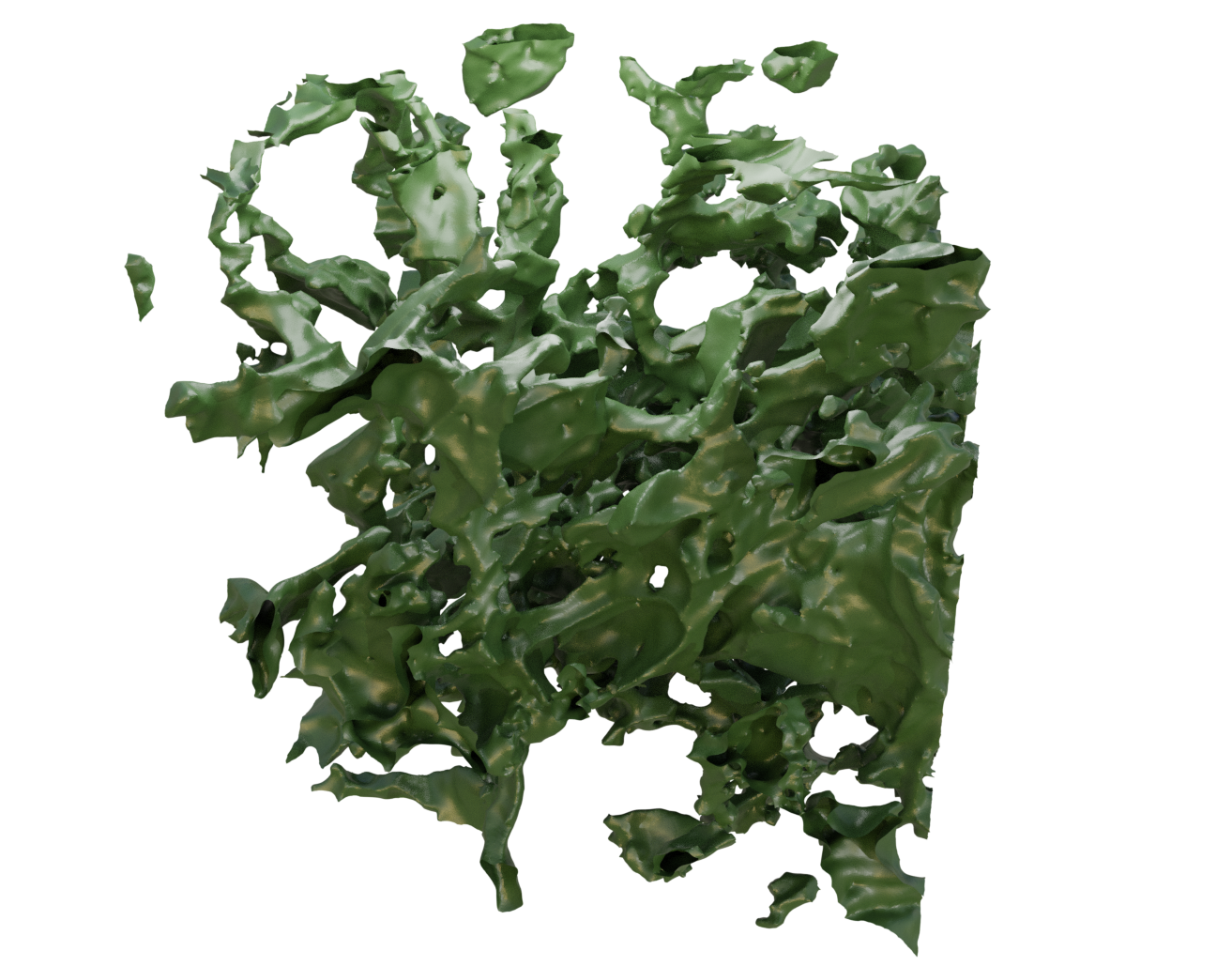}

    \caption{\textbf{Mice glia datasets:} full 3D reconstruction of the glia cells from the 3 stacks considered in this project~\cite{cali2018effects}.From left to right: Mouse 1, Mouse 3, and Mouse 4. }
    \label{fig:mice-glia}
\end{figure*}

We constructed a dataset from high-resolution 3D reconstructions of neuropil ultrastructure in the somatosensory cortex of mice. The dataset incorporates dense reconstructions obtained from serial electron microscopy (EM) images, focusing on axons, dendrites, synapses, and mitochondria, as detailed in~\cite{cali2018effects}. The source data was curated from the publicly available Dryad repository~\cite{cali2018effects}.

The dataset comprises three volumetric samples of dimensions $5 \mu m \times 5 \mu m \times 5 \mu m$, extracted from layer 1 of the somatosensory cortex of mice. These samples, corresponding to three experimental groups (Mouse 1, Mouse 3, and Mouse 4), were selected based on their morphological diversity and relevance to our analysis. Each sample includes voxelized binary masks representing glia, mitochondria, and post-synaptic densities, enabling comprehensive morphological and connectivity analyses.

\begin{figure*}[ht]
    \centering
    \includegraphics[width=0.49\textwidth]{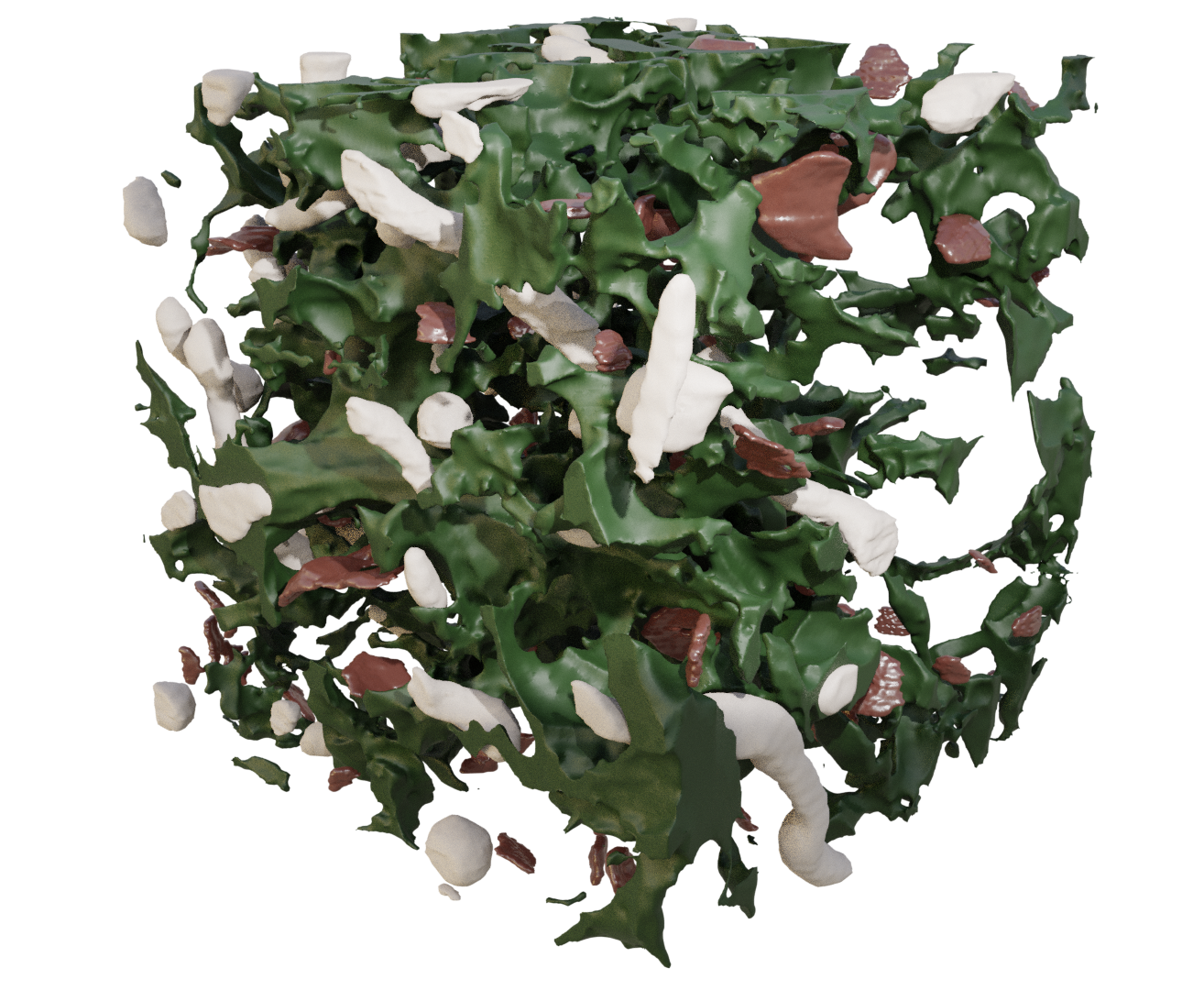}
    \includegraphics[width=0.49\textwidth]{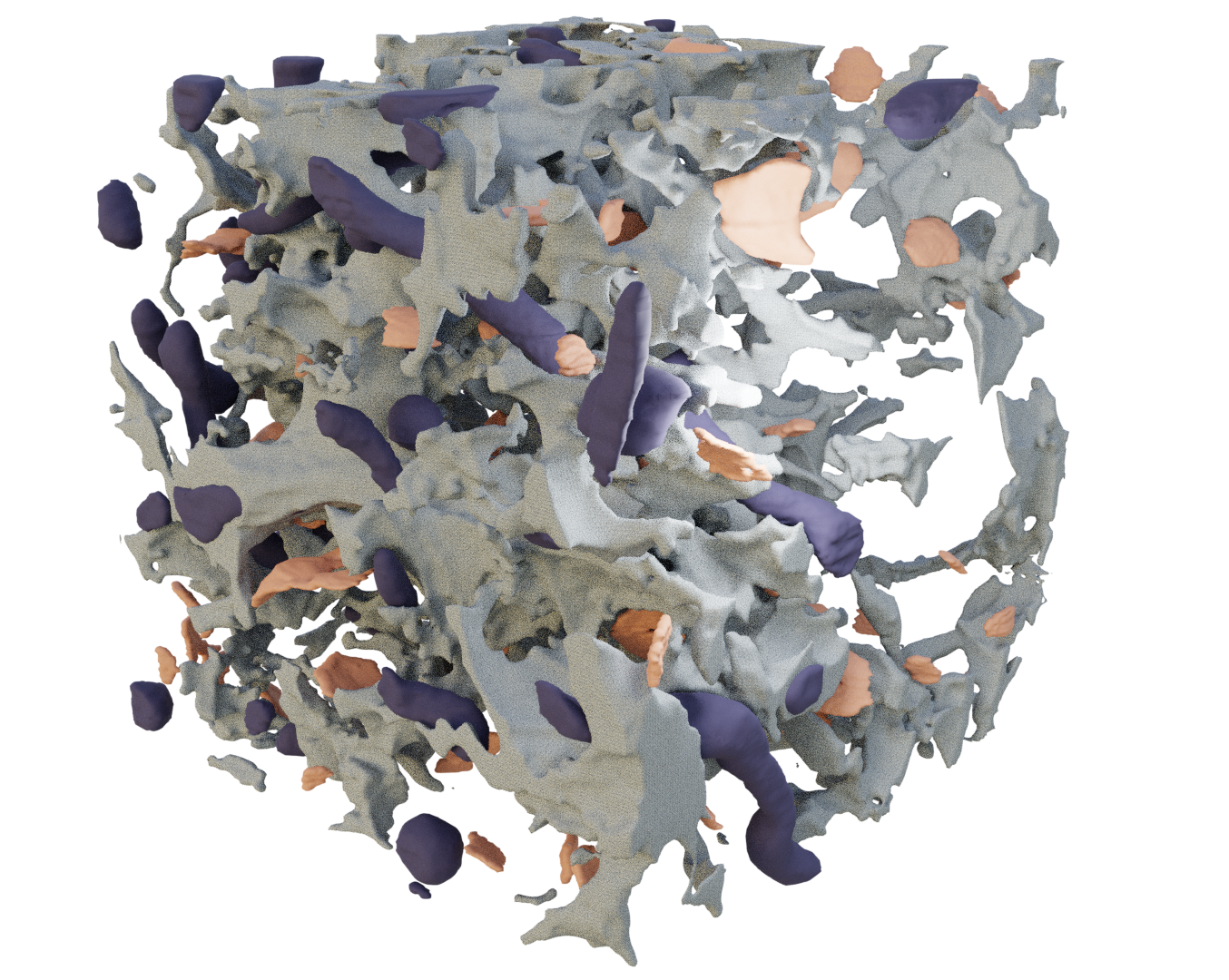}
    \caption{\textbf{Mouse 3 rendering:} Full 3D reconstruction of the data in the stack rendered using Blender software. Left: glia (green), mitochondria (white), synapses (pink). Right: glia (light gray), mitochondria (purple), and post-synaptic densities (orange).}
    \label{fig:mouse3-full-reconstruction}
\end{figure*}

To produce the segmentation masks, the high-resolution 3D reconstructions were processed using a combination of semi-automated segmentation tools (e.g., Ilastik and TrakEM2), followed by manual refinement for accuracy. These reconstructions were further voxelized to binary masks using custom scripts developed in Python, ensuring compatibility with downstream analyses. Each voxelized mask retains precise structural details, enabling quantitative analyses of cellular and synaptic elements.

We visualized and validated the dataset using Blender's advanced rendering capabilities to ensure the fidelity of the segmentation and the representational quality of the data. 
Fig.~\ref{fig:mice-glia} shows the full 3D reconstruction of glial cells from the three samples, highlighting the fractal complexity.
Fig.~\ref{fig:mouse3-full-reconstruction} illustrates the reconstructed glia, mitochondria, and post-synaptic densities from Mouse 3, highlighting the spatial organization and structural relationships within the sample.

The finalized dataset is publicly available via Dropbox (\url{https://bit.ly/42k3B1c}) and upon request for reproducibility and future research applications.

\section{Additional comparisons}
\label{sec:vit}

To comprehensively evaluate our method's performance, we conducted additional benchmarking against recent state-of-the-art transformer architectures in medical image segmentation. While several notable architectures such as ATFormer~\cite{Pan:2023:ATFormer}, DualRel~\cite{Mai:2023:DualRel}, and FragViT~\cite{Luo:2024:frag} show promising results in literature, their implementations were not publicly available for direct comparison. Therefore, we focused our comparison on two widely-adopted transformer-based architectures: TransUNet~\cite{chen2024transunet} and SwinUNet~\cite{cao2022swinunet}, which have demonstrated comparable performance to the aforementioned methods.

Given the architectural differences and convergence characteristics, we adapted the training protocols accordingly. SwinUNet, being a lightweight architecture, required extended training periods to achieve optimal performance. Specifically, we trained SwinUNet for 350 epochs on the Lucchi dataset and 150 epochs on other datasets, while TransUNet training was capped at 100 epochs across all datasets.

Table~\ref{tab:model-comparison} presents the quantitative results of this comparative analysis. TransUNet exhibited competitive performance across all datasets, achieving mIoU scores of 56.5\%, 70.5\%, 37.3\%, and 84.8\% on Mice-Glia, Mice-Mito, Mice-Synapses, and Lucchi datasets respectively, which are comparable to our proposed SAM4EM. SwinUNet, while being a lightweight architecture, showed relatively lower performance with mIoU scores of 48.4\%, 64.1\%, 32.9\%, and 78.1\% across the same datasets. These results demonstrate that while transformer-based architectures can achieve competitive performance in medical image segmentation, our proposed method maintains its advantages in terms of efficiency and consistency across diverse anatomical structures.

\begin{table*}[t]
\centering
\caption{Quantitative comparison across datasets using Dice coefficient (Dice$\uparrow$) and mean Intersection over Union (mIoU$\uparrow$) metrics. Higher values indicate better performance. Comparing the performance with recent transformer architectures  designed for medical image segmentation.}
\begin{tabular}{l|cc|cc|cc|cc}
\hline
\multirow{2}{*}{\textbf{Model}} 
    & \multicolumn{2}{c|}{\textbf{Mice-Glia}} 
    & \multicolumn{2}{c|}{\textbf{Mice-Mito}} 
    & \multicolumn{2}{c|}{\textbf{Mice-Synapses}} 
    & \multicolumn{2}{c}{\textbf{Lucchi}} \\
\cline{2-9}
& Dice$\uparrow$ & mIoU$\uparrow$ 
& Dice$\uparrow$ & mIoU$\uparrow$ 
& Dice$\uparrow$ & mIoU$\uparrow$ 
& Dice$\uparrow$ & mIoU$\uparrow$ \\
\hline
TransUNet \cite{chen2024transunet} 
    & 71.8 & 56.5 
    & 81.7 & 70.5 
    & 53.6 & 37.3 
    & 91.7 & 84.8 \\
SwinUNet \cite{cao2022swinunet} 
    & 63.7 & 48.4 
    & 76.8 & 64.1 
    & 46.3 & 32.9 
    & 85.7 & 78.1 \\
\hline
\end{tabular}
\label{tab:model-comparison}
\end{table*}

\begin{figure*}[ht]
    \centering
    \includegraphics[width=0.49\textwidth]{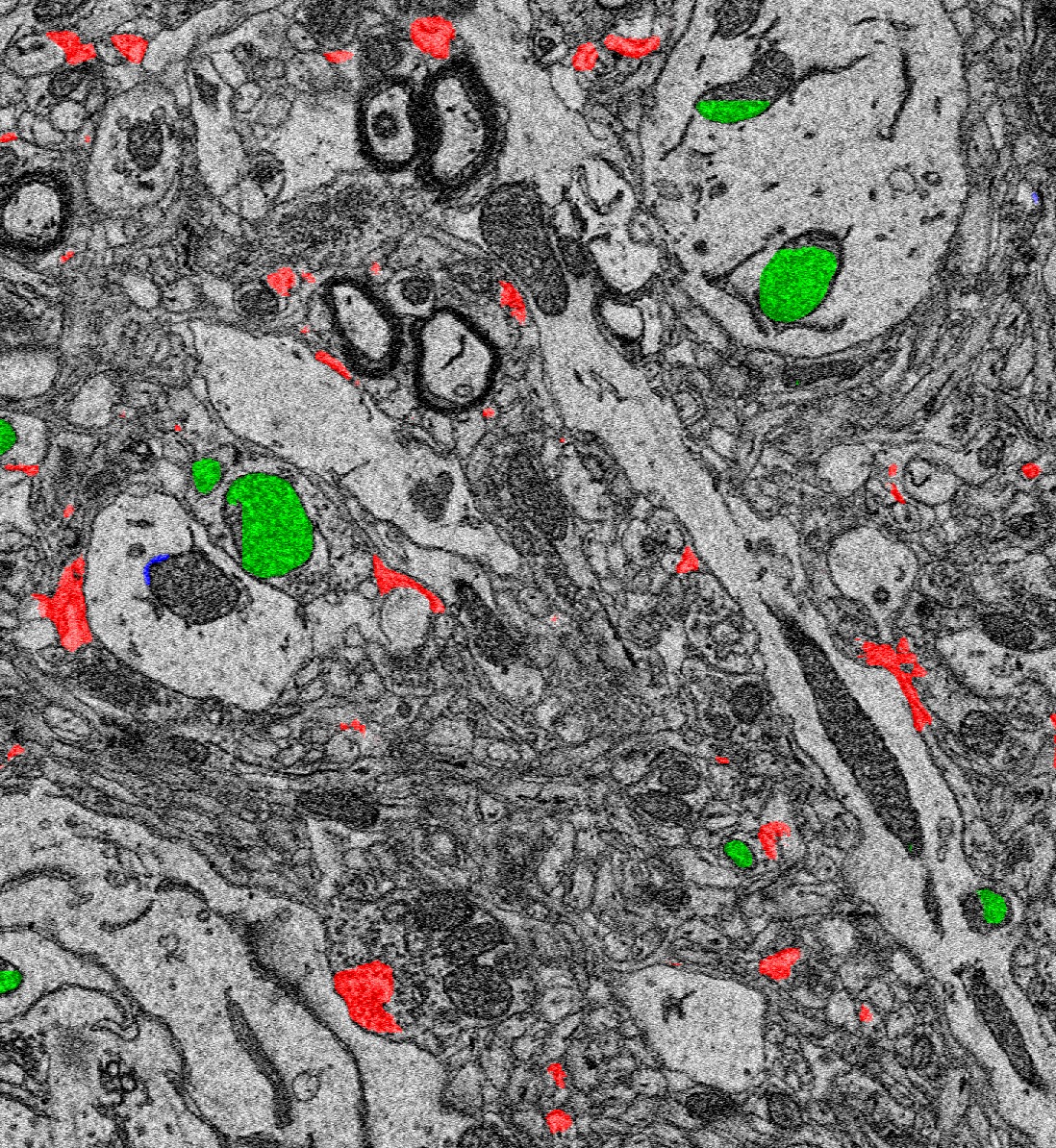}
    \includegraphics[width=0.49\textwidth]{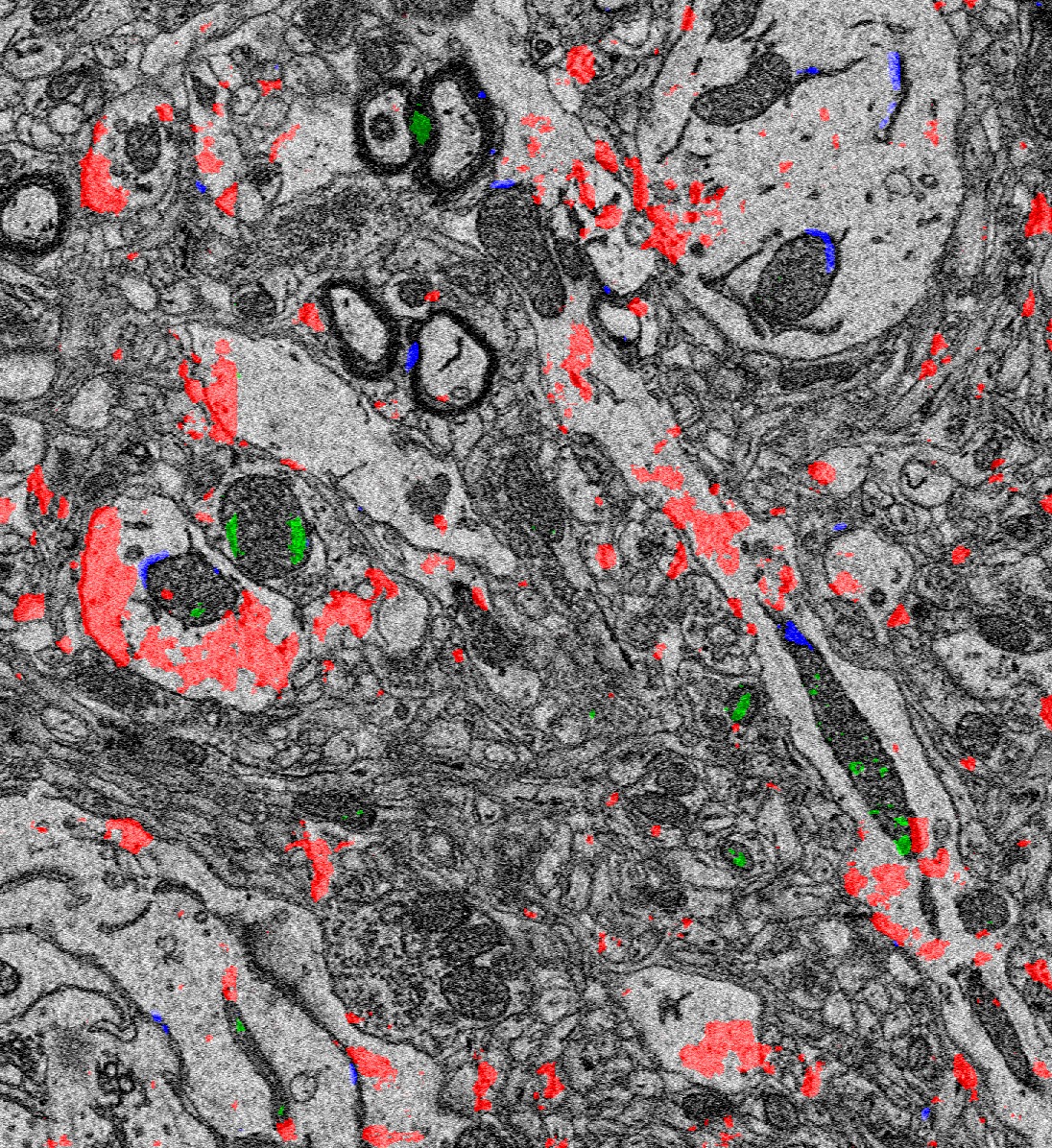}
    \includegraphics[width=0.47\textwidth]{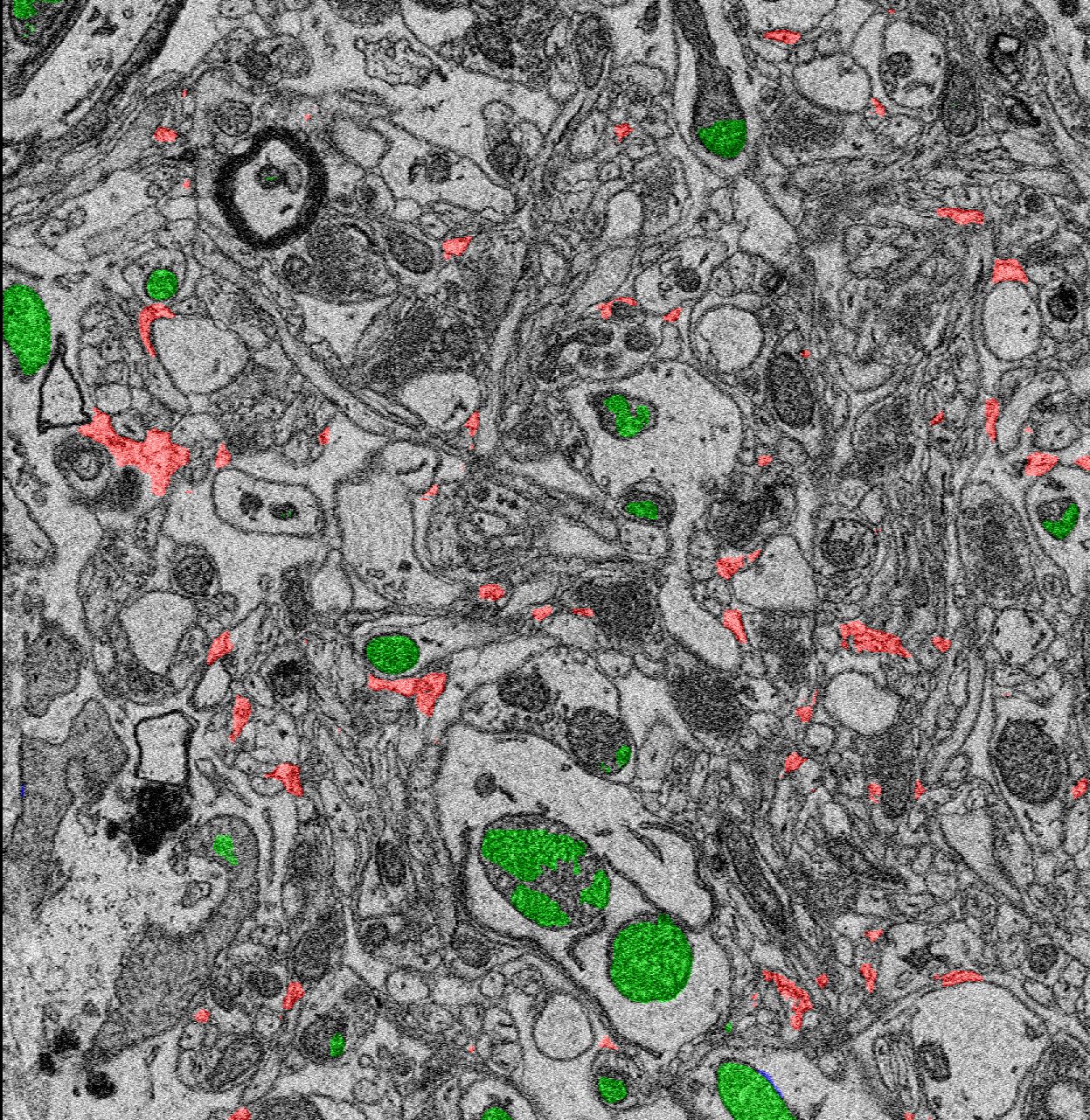}
    \includegraphics[width=0.50\textwidth]{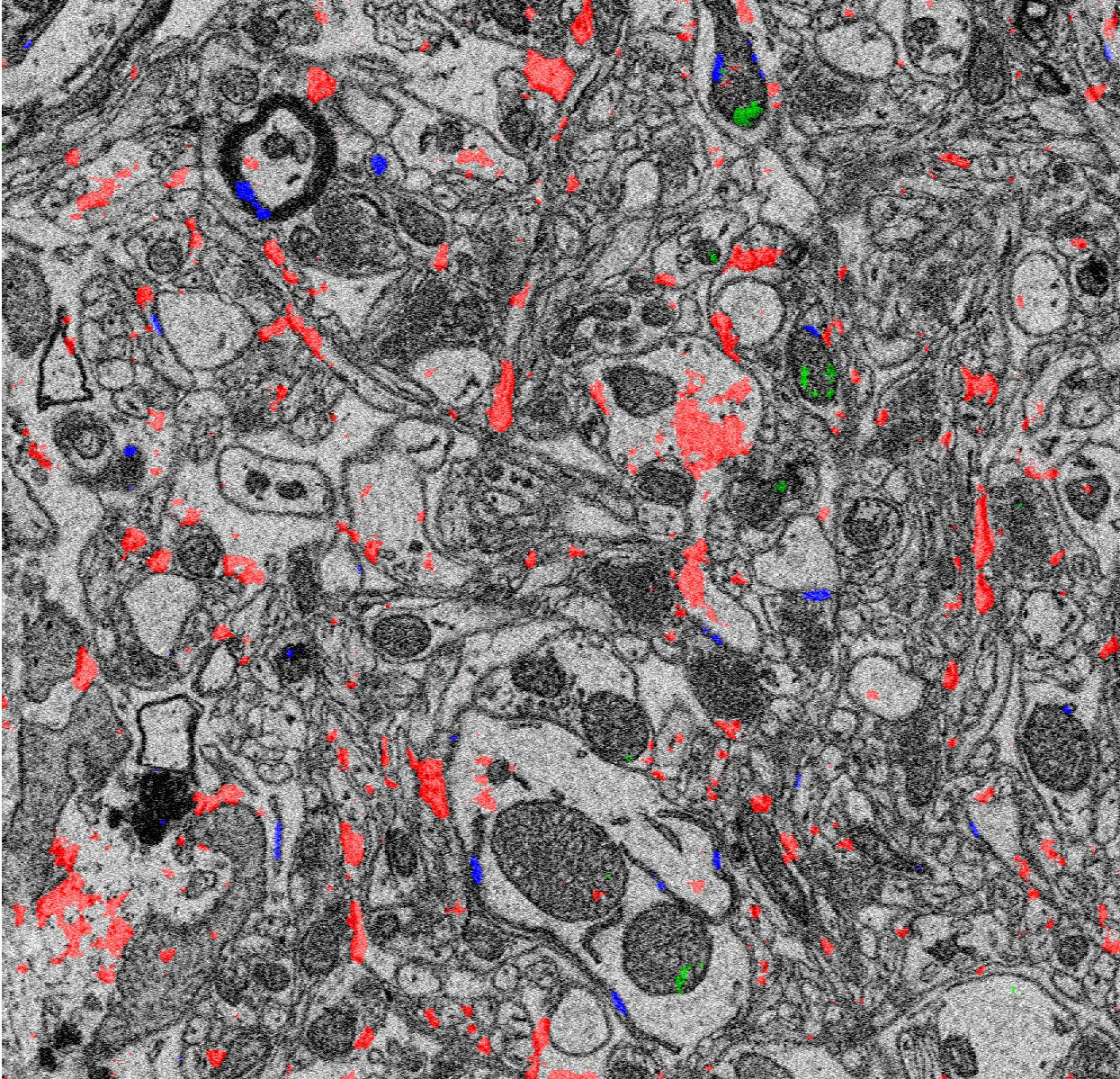}
    
    \caption{\textbf{Qualitative comparison:} a domain expert benchmarked the prpoosed SAM4EM and TransUnet~\cite{chen2024transunet} architecture on a stack representing brain parenchyma. From top to botto, two details are shown: on the left, the inference obtained with SAM4EM, on the right the output from TransUnet.}
    \label{fig:qual}
\end{figure*}

\section{Qualitative analysis}
\label{sec:qual}
To further challenge our method, we have tested SAM4EM on another EM dataset, acquired using different conditions. Specifically, this dataset was acquired using Serial block-face Electron Microscopy (SBF-EM), whose z-resolution is lower (50 nm-thick sections) compared to the training dataset, acquired with FIB-SEM (12 to 15nm thickness), but the system allows for larger fields of view. In particular we selected a stack from brain parenchyma, containing a cell body, of 50 micrometers width, at very high resolution (10 nm pixel size). The field of view of the training dataset was an order of magnitude smaller (5 micrometers, at 5 nanometers per pixel). Because of the different acquisition methods, images obtained using SBF-EM are qualitatively different, with a slightly higher noise.
A domain expert performed a qualitative assessment of the inference on this stack with our proposed SAM4EM against the transformer-based TransUNet~\cite{chen2024transunet}. Fig.~\ref{fig:qual} shows two details extracted from few slices: in red, the glia, in green the mitochondria, and in blue the post-synaptic densities. The  domain expert reported the following:
\begin{itemize}
\item The classifications obtained with both nets are rather poor, with a lot of objects missing per each category. This might be due, as previously mentioned, by the technical differences of the acquisition methods used to acquire the training datasets. 
\item Nevertheless, we can say that SAM4EM performed better than TransUNet. In fact, despite the paucity of objects detected, SAM4EM was still able to correctly detect mitochondria, although not all, and the red profile correspond to astrocytic processes. While TransUNet misclassified many of the processes detected, notably with mitochondria, which is basically non-existing, and the detection of glial processes included many neuronal processes. 
\item Finally both models performed extremely poorly on classification of synaptic densities. Most likely, by training SAM4EM with another, more similar dataset, it is likely that the segmentation might improve significantly, while TransUNet is visibly too faulty.
\end{itemize}

\section{Additional Discussions and Clarifications}

\subsection{Memory Efficiency and Scalability}

SAM4EM is designed to handle large-scale EM data efficiently during both training and inference. Our model achieves memory efficiency through several key mechanisms:
\begin{itemize}
    \item \textbf{LoRA adaptation:} Reduces trainable parameters by approximately 85\% compared to full fine-tuning
    \item \textbf{Efficient memory attention:} Utilizes an 8-slot attention mechanism instead of full self-attention
    \item \textbf{Sliding window processing:} Enables handling of large volumes during inference
\end{itemize}
In inference mode, SAM4EM successfully processes volumes up to $50μm$ width (approximately 10× larger than training volumes), as demonstrated in our qualitative analysis in Section 4. For context, the model requires approximately 4GB GPU memory for processing 512×512 resolution images, with peak usage not exceeding 6GB.

The method scales linearly with volume size through sliding window processing, maintaining consistent segmentation quality even on larger datasets. This efficiency enables deployment on moderate GPU hardware while retaining high segmentation accuracy on complex neural structures.

\subsection{Architectural Design Choices}
The architectural components of SAM4EM were specifically designed to address the challenges of segmenting complex fractal structures in EM data:
\begin{itemize}
    \item \textbf{Multi-scale feature enhancement:} Processes features at 1/4, 1/8, and 1/16 resolutions to effectively capture both fine details and broader contextual information critical for tracing intricate cellular processes
\item \textbf{3D memory-based attention:} Ensures structural consistency across consecutive slices, particularly important for maintaining coherence in branching cellular structures
\item \textbf{Bi-directional refinement mechanism:} Specifically targets irregular boundaries characteristic of glia cells and synaptic junctions
\end{itemize}
The effectiveness of these design choices is validated by our performance improvements on challenging structures, particularly glia cells (70.5\% vs. H-SAM's 68.7\% Dice) and synaptic junctions (53.8\% vs. H-SAM's 42.3\% Dice).

\subsection{Comparison Methodology}

Our evaluation focused on comparing SAM4EM with recent foundation model adaptations (H-SAM, SAMed, UN-SAM) rather than methods requiring full training from scratch. This decision was motivated by our research objective: investigating the potential of fine-tuning foundation models for complex 3D segmentation tasks with limited annotated data.

Traditional segmentation methods would require extensive labeled data for training, whereas our approach leverages the pre-trained knowledge of the SAM foundation model, achieving superior performance through efficient adaptation. This comparison framework better reflects the practical scenario where annotation resources are limited but high segmentation accuracy is required for complex neural structures.

\subsection{Future Work}

While this work focused on demonstrating SAM4EM's effectiveness on fully annotated datasets of moderate size, we recognize the potential for further reducing annotation requirements. Preliminary investigations suggest that our approach could significantly reduce the annotation burden through semi-supervised learning and active learning strategies.

Future work will explore:
\begin{itemize}
    \item Quantifying annotation time savings through interactive segmentation approaches
    \item Extending the method to handle dense segmentation tasks with multiple cell types and organelles
    \item Developing specialized data augmentation techniques for electron microscopy data to further reduce annotation requirements
\end{itemize} \\

%%%%%%%%% BODY TEXT
{\small
\textbf{Acknowledgments.}
This publication was funded by the PPM-7th Cycle grant (PPM 07-0409-240041, AMAL-For-Qatar) from the Qatar National Research Fund, a member of the Qatar Foundation. The findings herein reflect the
work and are solely the responsibility, of the authors.
}

%------------------------------------------------------------------------

%%%%%%%%% REFERENCES
\clearpage
{\small
\bibliographystyle{ieee_fullname}
\bibliography{sam-em}
}